\documentclass[sw,crcready]{iosart2x}

\usepackage{multirow}
\usepackage{soul}
\usepackage{pifont}% http://ctan.org/pkg/pifont
\newcommand{\cmark}{\ding{51}}%
\newcommand{\xmark}{\ding{55}}%

\usepackage[dvipsnames,table,xcdraw]{xcolor}
\hypersetup{
    colorlinks = true,
    linkcolor = blue,
    citecolor = black
}

\begin{document}

\begin{frontmatter}

%\pretitle{}
\title{Adversarial Transformer Language Models for Contextual Commonsense Inference}
\runtitle{Adversarial Transformer Language Models for Contextual Commonsense Inference}
\runauthor{Colon-Hernandez, et al.}
%\subtitle{}

% For one author:
%\author{\inits{N.}\fnms{Name1} \snm{Surname1}\ead[label=e1]{first@somewhere.com}}
%\address{Department first, \orgname{University or Company name},
%Abbreviate US states, \cny{Country}\printead[presep={\\}]{e1}}

% Two or more authors:
\begin{aug}
% \author{Anonymous Authors}
\author[A]{Pedro Colon-Hernandez}{pe25171@mit.edu}
\author[B]{Henry Lieberman}{lieber@media.mit.edu}
\author[C]{Yida Xin}{yxin@bu.edu}
\author[E]{Claire Yin}{yinc@mit.edu}
\author[A]{Cynthia Breazeal}{cynthiab@media.mit.edu}
\author[D]{Peter Chin}{spchin@cs.bu.edu}
\address[A]{Media Lab, \orgname{Massachusetts Institute of Technology}}
\address[B]{CSAIL, \orgname{Massachusetts Institute of Technology}}
\address[C]{Work done while at Department of Computer Science \orgname{Boston University}}
\address[D]{Center for Brains, Minds and Machines MIT \& Thayer School of Engineering, Dartmouth College}
\address[E]{Work done while at CSAIL, \orgname{Massachusetts Institute of Technology}}

% \author[A]{\inits{N.}\fnms{Name1} \snm{Surname1}\ead[label=e1]{first@somewhere.com}%
% \thanks{Corresponding author. \printead{e1}.}}
% \author[B]{\inits{N.N.}\fnms{Name2 Name2} \snm{Surname2}\ead[label=e2]{second@somewhere.com}}
% \author[B]{\inits{N.-N.}\fnms{Name3-Name3} \snm{Surname3}\ead[label=e3]{third@somewhere.com}}
% \address[A]{Department first, \orgname{University or Company name},
% Abbreviate US states, \cny{Country}\printead[presep={\\}]{e1}}
% \address[B]{Department first, \orgname{University or Company name},
% Abbreviate US states, \cny{Country}\printead[presep={\\}]{e2,e3}}
\end{aug}

%\begin{review}{editor}
%\reviewer{\fnms{First} \snm{Editor}\address{\orgname{University or Company name}, \cny{Country}}}
%\reviewer{\fnms{Second} \snm{Editor}\address{\orgname{First University or Company name}, \cny{Country}
%    and \orgname{Second University or Company name}, \cny{Country}}}
%\end{review}
%\begin{review}{solicited}
%\reviewer{\fnms{First} \snm{Solicited reviewer}\address{\orgname{University or Company name}, \cny{Country}}}
%\reviewer{\snm{anonymous reviewer}}
%\end{review}
%\begin{review}{open}
%\reviewer{\fnms{First} \snm{Open Reviewer}\address{\orgname{University or Company name}, \cny{Country}}}
%\end{review}

\begin{abstract}
\hyperref[def:contextualcommonsenseinference]{Contextualized or discourse aware commonsense inference}  \cite{Gabriel_PC} is the task of generating commonsense \hyperref[def:assertion]{assertions} (i.e., facts) from a given story, and a sentence from that story.  (Here, we think of a story as a sequence of causally-related events and descriptions of situations.) This task is hard, even for modern contextual language models. Some problems with the task are: lack of controllability for topics of the inferred \hyperref[def:assertion]{assertions}; lack of commonsense knowledge during pre-training; and, possibly, hallucinated or false \hyperref[def:assertion]{assertions}. The task's goals are to make sure that (1) the generated \hyperref[def:assertion]{assertions} are plausible as commonsense; and (2) to assure that they are appropriate to the particular context of the story.

We utilize a transformer model as a base inference engine to infer commonsense \hyperref[def:assertion]{assertions} from a sentence within the context of a story.  With our inference engine we address lack of controllability, lack of sufficient commonsense knowledge, and plausibility of assertions through three techniques. We control the inference by introducing a new technique we call “hinting”. \hyperref[def:hinting]{Hinting} is a kind of language model prompting \cite{liu2021pre}, that utilizes both hard prompts (specific words) and soft prompts (virtual learnable templates). This serves as a control signal to advise the language model “what to talk about”. Next, we establish a methodology for performing \hyperref[def:jointcommonsenseinference]{joint inference} with multiple commonsense knowledge bases.  While in logic, joint inference is just a matter of a conjunction of \hyperref[def:assertion]{assertions}, \hyperref[def:jointcommonsenseinference]{joint inference of commonsense} requires more care, because it is imprecise and the level of generality is more flexible. You want to be sure that the results “still make sense” for the context. To this end, we align the \hyperref[def:assertion]{assertions} in three \hyperref[def:knowledgegraph]{knowledge graphs }(ConceptNet \cite{speer2017conceptnet}, ATOMIC2020 \cite{hwang2020comet}, and GLUCOSE \cite{mostafazadeh-etal-2020-GLUCOSE}) with a story and a target sentence, and replace their symbolic \hyperref[def:assertion]{assertions} with textual versions of them. This combination allows us to train a single model to perform \hyperref[def:jointcommonsenseinference]{joint inference} with multiple \hyperref[def:knowledgegraph]{knowledge graphs}. We show experimental results for the three \hyperref[def:knowledgegraph]{knowledge graphs }on \hyperref[def:jointcommonsenseinference]{joint inference}. Our final contribution is a \hyperref[def:gan]{GAN} architecture that generates the \hyperref[def:contextualcommonsenseinference]{contextualized commonsense inference} from stories and scores the generated \hyperref[def:assertion]{assertions} as to their plausibility through a discriminator.  The result is an integrated system for \hyperref[def:contextualcommonsenseinference]{contextual commonsense inference} in stories, that can controllably generate plausible commonsense \hyperref[def:assertion]{assertions}, and takes advantage of \hyperref[def:jointcommonsenseinference]{joint inference} between multiple commonsense knowledge bases.
\end{abstract}


\end{frontmatter}

%%%%%%%%%%% The article body starts:

\section{Introduction}
\begin{figure}[ht!]
\includegraphics[width=8cm]{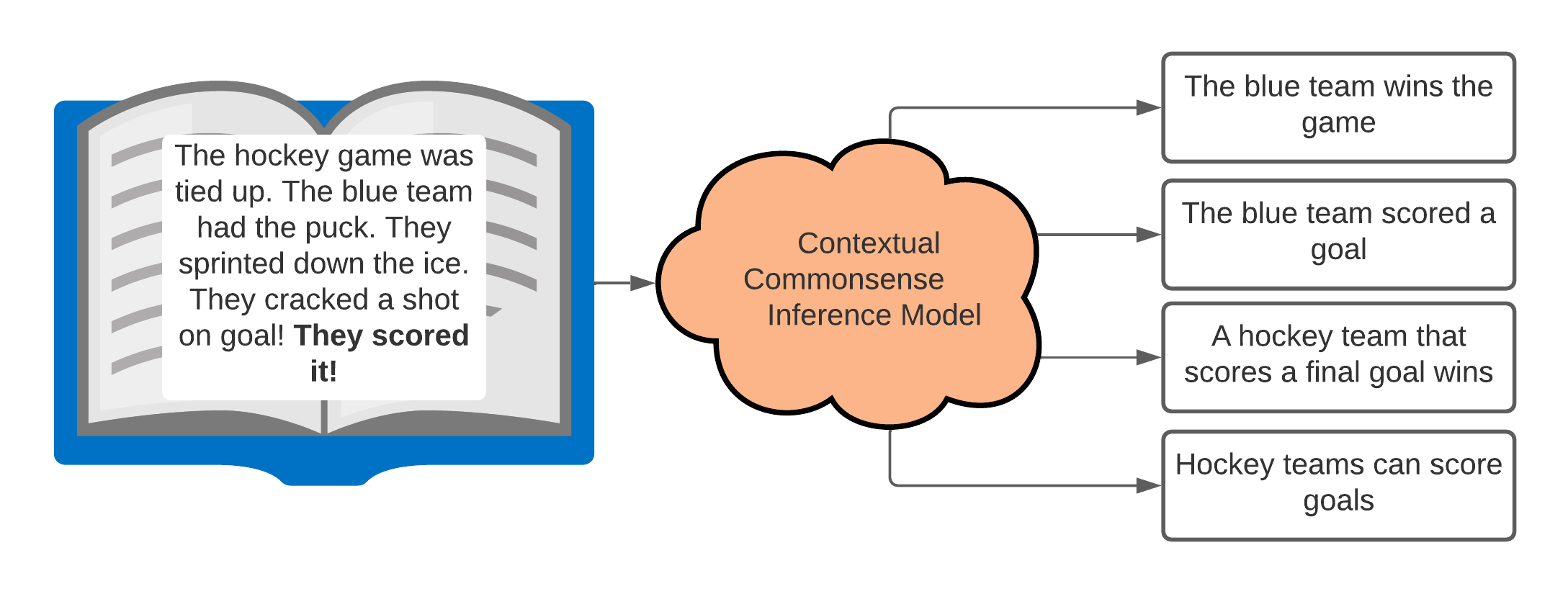}\caption{Overview of the task of \hyperref[def:contextualcommonsenseinference]{contextual commonsense inference}. From the story on the left, and the \textbf{bolded} sentence, a model should infer \hyperref[def:assertion]{assertions} such as the ones on the right.}\label{fig:contcommonsenseinf}
\end{figure}

\hyperref[def:contextualcommonsenseinference]{Contextualized or discourse aware commonsense inference} \cite{Gabriel_PC} is a task in which we are given a text context (e.g., story) and a selected sentence from that context, and we have to infer a coherent and contextual \hyperref[def:assertion]{commonsense assertion (i.e., fact)} from the given context and target sentence. We extend this definition to additionally include \hyperref[def:storyspecificinference]{story specific assertion inferences} (i.e., templates that are instanced by elements from a story), or \hyperref[def:generalinference]{general assertion inferences} (i.e., fact templates) as used in \cite{mostafazadeh-etal-2020-GLUCOSE}. This framing of a text context and target sentence is important because the text could be a story, a procedure, etc. We define an \textit{assertion} here as a tuple that represents a fact.  This tuple contains at least a \textcolor{Emerald}{subject}, \textcolor{Orchid}{a relation type}, and \textcolor{Peach}{an object} (similar to subject-verb-object triples).  We add a field to this tuple, which is \textit{specificity}.  We define \hyperref[def:specificity]{\textit{specificity}} as whether the assertion's content is about entities in the aligned story, or if it is a generalized version of an assertion. This can be seen as whether the assertion is a \hyperref[def:generalinference]{\textit{general}} template with variables, or a \hyperref[def:storyspecificinference]{\textit{specific}} instance of this template.   In the case of a story, \hyperref[def:contextualcommonsenseinference]{contextual commonsense inference} can help with story understanding (e.g., a \hyperref[def:contextualcommonsenseinference]{contextual commonsense inference} system could infer \hyperref[def:assertion]{assertions} in a story that indicate a revenge plot \cite{williams2017commonsense}), and in the case of a procedure, it could help with step explanations and step rephrasing by giving possibly unstated \hyperref[def:assertion]{assertions}. Such framing additionally allows us to utilize a trained \hyperref[def:contextualcommonsenseinference]{contextual commonsense inference} model downstream by going sentence-by-sentence, inferring \hyperref[def:assertion]{assertions} as the context changes. We give an example of the task in Figure \ref{fig:contcommonsenseinf}. 

To clarify the task of \hyperref[def:contextualcommonsenseinference]{contextual commonsense inference} even further, below we give an example with a story, \textbf{a target sentence}, and some corresponding \hyperref[def:storyspecificinference]{story specific} and \hyperref[def:generalinference]{general} commonsense assertion inferences. The story comes directly from the ROCStories corpus \cite{mostafazadeh2016corpus}.  

\begin{quote}
\small
\textbf{Story:} The hockey game was tied up. The \textcolor{red}{red team} had the puck. They sprinted down the ice. They cracked a shot on goal! \textbf{They scored a final goal!}\\

\textbf{Story Specific Commonsense Inference:} \textcolor{Emerald}{The {\setulcolor{red}\ul{red team}}}, \textcolor{Orchid}{is capable of}, \textcolor{Peach}{winning the game}\\

\textbf{General Commonsense Inference:} \textcolor{Emerald}{ {\setulcolor{red}\ul{Some people}} \textit{scored a final goal} }, \textcolor{Orchid}{causes}, \textcolor{Peach}{ {\setulcolor{red}\ul{some people}} to be happy}
\end{quote}

This task is hard for modern pre-trained contextual language models \cite{Gabriel_PC}. This may be because it may rely on information that a model may not have seen during pre-training, or the model has to figure out what topic to infer information about. The first issue is exacerbated because commonsense knowledge, present in everyone and target of a model trained in this task, tends to not be written explicitly in text \cite{zhang2021knowledge,liu2004conceptnet,da2019cracking,davison2019commonsense}. In addition to these problems, the correctness of the information that is generated by the models is hard to evaluate and usually involves a costly human-in-the-loop setup. 

Prior work, such as COMET \cite{bosselut2019comet}, has tried to do  \hyperref[def:sentencecommonsenseinference]{sentence-level commonsense inference}: generating a commonsense \hyperref[def:assertion]{assertion}, with at most a sentence as context.   ParaCOMET \cite{Gabriel_PC} is an extension of COMET that was developed to work at a paragraph-level (i.e., what we describe as the \hyperref[def:contextualcommonsenseinference]{contextual commonsense inference} task). ParaCOMET utilizes a recurrent memory and is trained on a corpus of aligned stories and \hyperref[def:assertion]{assertions}. ParaCOMET builds this dataset to address the \hyperref[def:contextualcommonsenseinference]{contextual commonsense inference} task by aligning facts from a commonsense \hyperref[def:knowledgegraph]{knowledge graph }(i.e., ATOMIC \cite{sap2019ATOMIC}) with a story (i.e., sampled from ROCStories \cite{mostafazadeh-etal-2016-corpus}) through a heuristic based on the ROUGE \cite{lin-2004-rouge} metric.  It goes a step further by utilizing the cross entropy of story tokens of a language model, conditioned on one of the matched facts, as a measure of coherence to keep only \hyperref[def:assertion]{assertion} matches that are coherent to the narrative. They additionally address the need for memory (i.e., for the model to remember prior events) by using and saving prior aligned \hyperref[def:assertion]{assertions} in a memory system. An example of an input and expected output from ParaCOMET can be seen below:
\begin{quote}
\small
\textbf{Model Input:} The hockey game was tied up. The \textcolor{red}{red team} had the puck. They sprinted down the ice. They cracked a shot on goal! \textbf{They scored a final goal!} <|sent5|> <|xEffect|>\\

\textbf{Model Target/Output:}  \textcolor{Peach}{win the game}\\
\end{quote}

In this example, since the model is predicting ATOMIC \textcolor{Peach}{objects}, the output is a single phrase (i.e., \textcolor{Peach}{win the game}). Additionally, the symbols \textit{<|sent5|>} and \textit{<|xEffect|>} mean that the target sentence is sentence number five\footnote{We note that in the original ParaCOMET work, the sentences were 0-start indexed. We utilize 1-start indexing for clearer understanding.}, and that the relation we want to generate a tuple about is the “has the effect on a certain person(s)” respectively.  

Another parallel work that has tackled \hyperref[def:contextualcommonsenseinference]{contextual commonsense inference} is GLUCOSE \cite{mostafazadeh-etal-2020-GLUCOSE}.  GLUCOSE annotates the ROCStories \cite{mostafazadeh-etal-2016-corpus} corpus along ten dimensions of commonsense.  The authors annotate every sentence with an \hyperref[def:assertion]{assertion} that is either present or implied in it for a given dimension. Additionally, they annotate each assertion with a general version of it, as we defined previously as \hyperref[def:generalinference]{general inferences}, which includes variables and their descriptions. What this means is that any person(s) or object(s) in the \hyperref[def:assertion]{assertion} is(are) replaced with a token such as \textit{Person\_A}, etc. to represent a “general” version of the fact. An example of the GLUCOSE formulation's inputs and expected outputs is given below:

\begin{quote}
\small
\textbf{Model Input:} 1: The hockey game was tied up. The \textcolor{red}{red team} had the puck. They sprinted down the ice. They cracked a shot on goal! *\textbf{They scored a final goal!}*\\

\textbf{Model Target/Output:} \textcolor{Emerald}{The red team scores}, \textcolor{Orchid}{Causes/Enables}, \textcolor{Peach}{they win the game} ** \textcolor{Emerald}{People\_A score}, \textcolor{Orchid}{Causes/Enables}, \textcolor{Peach}{People\_A win a game}
\end{quote}

This formulation of \hyperref[def:contextualcommonsenseinference]{contextual commonsense inference} is harder than the ParaCOMET one because it has to generate two sets of \hyperref[def:assertion]{assertions}  \textcolor{Emerald}{a subject}, \textcolor{Orchid}{relation}, and \textcolor{Peach}{an object} tuples, where one is the \hyperref[def:storyspecificinference]{story specific} one and the other is the \hyperref[def:generalinference]{general version} of the assertion. These are seen above, separated by the ** respectively. In this example additionally, we can see the symbol \textit{1:} which tells the model to predict along a dimension of commonsense described by GLUCOSE (i.e., 1: Event that directly causes or enables X), and the sentence enclosed by asterisks (*) which signifies it is the target sentence. With this corpus of annotated stories, the authors train a T5 \cite{2020t5} model to, given a dimension, target sentence, and story, generate both a \hyperref[def:storyspecificinference]{story specific} and \hyperref[def:generalinference]{general}  \hyperref[def:assertion]{assertion} .  It is worth noting that in both works, the models have to do their inference from the story, a target sentence, and relation alone.

However, none of these works address controllability in the generation, which means that the models can generate \hyperref[def:assertion]{assertions} that may be irrelevant to the sentence, or may not be about a topic needed for a downstream application. Additionally, these models are only trained on one dataset at a time, which can hinder a model's capability to infer knowledge if it  has not seen the knowledge elsewhere. Lastly, these models do not score the factuality or correctness of the \hyperref[def:assertion]{assertions}; at most they can generate a beam score, which indicates the likelihood of the generated phrase. 

In this work, we attempt to address all of these shortcomings through various techniques. Firstly, we construct a dataset of contextualized \hyperref[def:assertion]{assertions} consisting of \hyperref[def:assertion]{assertions} from ConceptNet \cite{speer2017conceptnet}, ATOMIC 2020 \cite{hwang2020comet}, and GLUCOSE \cite{mostafazadeh-etal-2020-GLUCOSE}. To construct this dataset, we align the ROCStories \cite{mostafazadeh-etal-2016-corpus} with an \hyperref[def:assertion]{assertion} by generating sentence/paragraph embeddings for the stories and the \hyperref[def:assertion]{assertions} by using the sentence-transformers
 \cite{reimers-2019-sentence-bert} library. We then use cosine distance to find the closest story for each \hyperref[def:assertion]{assertion}. With this closest story, we repeat the process once more, now with the sentences from the story, to find the closest sentence in the story to the \hyperref[def:assertion]{assertion}.  This contextualization and alignment puts all the knowledge bases in the same contextual universe. Secondly, we augment this dataset of aligned \hyperref[def:assertion]{assertions}, stories, and target sentences, with \hyperref[def:hint]{“hints”}, as a method to communicate constraints when performing contextual commonsense inference.  We automatically generate \hyperref[def:hint]{“hints”} by selecting parts of a target assertion, along with a symbol identifying the parts. Lastly, we use this dataset to adversarially train two language models; one to infer \hyperref[def:assertion]{assertions} from the story and target sentence, and a second to validate or score the \hyperref[def:assertion]{assertion} given the story and the target sentence.  This adversarial set of models can be utilized in downstream applications for tasks such as \hyperref[def:knowledgegraph]{knowledge graph }construction, and leverages multiple knowledge sources to infer and score commonsense inferences.   

Altogether, our contributions in this work are:
\begin{itemize}
    \item The utilization of a hinting mechanism to help condition and control a generative model for \hyperref[def:contextualcommonsenseinference]{contextual commonsense inference}. 
    \item A simple method for contextualizing \hyperref[def:assertion]{assertions} to a given text with the purpose of performing \hyperref[def:jointcommonsenseinference]{joint inference}. 
    \item A method for adversarially training language models to infer and evaluate \hyperref[def:assertion]{assertions} from a story context.
    
\end{itemize}

This work is organized in the following manner. We begin with a Related Work section, and follow it into the Hinting and Joint Inference techniques. Once we have laid out the groundwork for these, we finish with the GAN approach and a Conclusion. Throughout the work, we utilize various vocabulary items and compile these into a Definitions appendix, along with links to a corresponding definition. 

\section{Related Work}

\subsection{Prompting}
\label{sec:prompting}
Recently, there has been a shift  in Natural Language Processing from pre-training and fine-tuning a model, to pre-training, prompting, and predicting \cite{liu2021pre}.  One  reason for this shift is the creation of ever-larger language models, which have become computationally expensive to fine-tune. Prompting is a finding a way to convert a model's input sequence into another sequence that resembles what the model has seen during pre-training. Overall, most prompting research focuses on formulating the task as a \textit{cloze} (fill-in-the-blanks) task.  However, we consider the task of language generation, an open-ended formulation. 

Recall that prefix prompting modifies the input to a language model, by adding either a hard prompt (additional words to the input sequence) \cite{shin2020autoprompt} or a soft prompt (i.e., adding trainable vectors that represent, but are not equivalent to, additional words) \cite{li2021prefix,lester2021power,liu2021pre}. Unlike classic prefix prompting, \hyperref[def:hinting]{hinting} uses both hard and soft prompts. The soft prompts are in the form of symbols that represent the different parts of the assertion (i.e., \textcolor{Emerald}{ subject} (\textit{<subj>}), \textcolor{Orchid}{relation type} (\textit{<relation>}), and \textcolor{Peach}{an object} (\textit{<obj>})), and the hard prompts are in the form of the actual parts of the assertion that are selected to be appended as part of the \hyperref[def:hint]{hint} as seen in our example in section \ref{sec:prompting}.  \hyperref[def:hinting]{Hinting} is similar to KnowPrompt \cite{DBLP:journals/corr/abs-2104-07650}, except that they use a masked language model and soft prompts for relationship extraction. AutoPrompt \cite{shin2020autoprompt} is also similar, but finds a set of “trigger” words that give the best performance on a \textit{cloze}-related task, whereas we provide specific structured input for the model to guide text generation.   

 We classify \hyperref[def:hinting]{hinting} as a hybrid prefix-prompting technique due to the inclusion of trainable symbols that are not part of the model's original vocabulary and can be viewed as soft-prompts given to the model, as well as the combination of these soft-prompts with actual hard prompts to generate a contextual inference. We refer to \cite{liu2021pre}'s definition of prompting as a three-step process. To begin, we define a function that converts the input to an intermediate template, more precisely in our case by including the hint between parenthesis at the end of the input. Second, the model must generate an answer $Z$ at the end of its input (hence the prefix prompting), which is context-dependent commonsense inference. Finally, this $Z$ is directly mapped to a $Y$, which corresponds to the target contextual \hyperref[def:assertion]{assertion} in the \hyperref[def:contextualcommonsenseinference]{contextual commonsense inference task}.

\subsection{Controllable Generation}
Controllable generation can be described as ways to control a language model's text generation given some kind of guidance.  One work that tries to implement controllable generation is CTRL \cite{keskar2019ctrl}.  The authors supply control signals during pre-training of a language model. This approach is intended to provide a generally applicable language model.  A body of work in controllable generation has focused on how it can be used for summarization.  Representative work that uses techniques similar to ours is GSum \cite{dou2020gsum}. In contrast to GSum, our method of \hyperref[def:hinting]{hinting} is model independent, allows for the source document to interact with the guidance signal, and contains soft prompts in the form of trainable embeddings that represent the parts of a tuple. The GSum system gives interesting insight into the fact that highlighted sentences, and the provision of triples, does in fact help with the factual correctness of abstractive summarization. 

We make the distinction that \hyperref[def:hinting]{hinting} falls more under prompting for the reason that we utilize additionally the trainable soft embeddings rather than purely additional hard tokens and that our task of contextual commonsense generation is not explored in the controllable generation works, whose main focus is on controlling unstructured text generation. Some works that are in this area are also \cite{peng2018towards} who utilize what they call "control factors" as keywords or phrases that are supplied by a human-in-the-loop to guide a conversation. More similar to our work, but tailored for the task of interactive story generation and without trainable soft-embeddings, is the work by \cite{brahman-etal-2020-cue} which uses automatically extracted keywords to generate a story. Future work we could possibly utilize the automatic keyword extraction to supply parts of a \hyperref[def:hint]{hint}, rather than our approach of complete parts of an \hyperref[def:assertion]{assertion}, and expand this to utilize synonyms of keywords. Lastly, there is the work by  \cite{see-etal-2019-makes} which looks at controllable text generation for the purpose of conversation and utilizes an embedding give quantitative control signals as part of conditional training.

\subsection{Story and Assertion Alignment}
The closest works to ours, with respect to constructing a story aligned assertion dataset, are ParaCOMET and GLUCOSE \cite{Gabriel_PC, mostafazadeh-etal-2020-GLUCOSE}.  GLUCOSE uses human annotation to perform the alignment between stories and commonsense \hyperref[def:assertion]{assertions}.  ParaCOMET takes an automated approach in which \hyperref[def:assertion]{assertions} are aligned either by giving the sentence to a COMET model as an input and producing a relevant inferred assertion, or by  calculating the cross entropy of combining the story up until the target sentence with an assertion from a knowledge base.  Our method differs from this in that we utilize cosine distance between semantic representations of the story and its sentences and an assertion from a knowledge base.  Some possible differences that arise from this is that our method could match \hyperref[def:assertion]{assertions} that may not be explicit in a story to that story.  Whereas ParaCOMET's approaches, which are based on cross-entropy for coherence, are likely to produce \hyperref[def:assertion]{assertions} that have parts that are explicit in text. Overall, our approach can match more abstract \hyperref[def:assertion]{assertions} to stories.  Additionally, our method permits us to use the optimized FAISS library to scale up to billions of stories and \hyperref[def:assertion]{assertions}, and gives us the freedom to select how to embed the stories/sentences/\hyperref[def:assertion]{assertions}.  

\subsection{Commonsense: Grounding, Reasoning, and Knowledge}
A related line of work  has been in grounding commonsense statements for inference.  However, this line of work is more aligned with natural language inference rather than \hyperref[def:assertion]{assertions}. One contribution in this area is HellaSwag \cite{zellers2019hellaswag} which constructs a question answering dataset whose plausible answers are intended to be confounders to language models. Our work differs from this line of work in that we intend to produce structured outputs. 

Other work looks at reasoning with commonsense knowledge graphs. One work that utilizes the explicit graph structure to perform multi-hop reasoning is “Commonsense for Generative Multi-Hop Question Answering Tasks” \cite{bauerwang2019commonsense}. The authors look to select grounded multi-hop relational commonsense information from ConceptNet via a pointwise mutual information and term-frequency based scoring function to fill in gaps of reasoning between context hops for a model they use. In contrast to this work, we are not explicitly looking at a graph structure for our inference, nor are looking at the task of question answering. 

Some older work that looks at doing something similar to \hyperref[def:jointcommonsenseinference]{joint inference} is blending \cite{5172887}.  This technique essentially consists of constructing and adding or blending together matrices of embeddings to find the commonalities between discrete knowledge sources and commonsense knowledge.  This method, however, is hard to scale to large knowledge bases, and is not easily applied to the task of contextual knowledge inference.  An even older project that looks into a certain kind of \hyperref[def:jointcommonsenseinference]{joint inference} is Cyc \cite{10.1145/79173.79176}. Cyc uses the idea of “micro-theories”; there would be a small set of commonsense \hyperref[def:assertion]{assertions} that you could reason with, then combine them with the more general Cyc KB. However, this is not really \hyperref[def:jointcommonsenseinference]{joint inference} in the sense that we use, but rather trying to address the problem of local vs. global inference. 

We examine also the work by \cite{chalier2020joint} which utilizes formal logic and restructuring of assertions to be able to combine and perform joint inference over knowledge bases, however the system as it is cannot be used for contextual commonsense inference because it requires explicit knowledge to be already present in a knowledge base to make inferences, whereas in our work through the underlying language models can produce inferences for unseen concepts. Additionally, we look at the work \cite{10.1145/3442381.3449827}, which uses a combination of systems to extract high quality non-triple formatted facts from text. However, the system is based on grammatical structures in an input text, which means that implied facts may not be extracted from this if such a system were utilized for contextual commonsense inference.

Other works that have tried to consolidate commonsense knowledge are the following. \cite{ILIEVSKI2021107347} examines multiple sources of knowledge and unifies the relations in these under 13 dimensions of commonsense, however it still remains a challenge to unify the nodes in the different sources, and such a broad unification may make it challenging to generate inferences for detailed relations (i.e., a specific relation type).

Lastly, we mention some works on open knowledge bases that could be leveraged in future work for utilization in joint inference: TupleKB \cite{mishra2017domain}, Quasimodo \cite{10.1145/3357384.3357955}, Ascent \cite{nguyen2021advanced}, GenericsKB \cite{bhakthavatsalam2020genericskb}.

\subsection{Adversarial Language Models}
\label{sec:ganlit}
Here we look into work that utilizes adversarial or pair training with language models.  One such work is \cite{west2021symbolic} in which the authors utilize a GPT-3 \cite{NEURIPS2020_1457c0d6} model as a teacher in order to distill commonsense knowledge into a student model that is considerably smaller. This task is different from ours in that they do not explore contextual/discourse aware commonsense inference, instead they look at extracting the knowledge already found in a model.

Other work more similar to ours, albeit older, is \cite{press2017language}.  In this work, the authors take a similar approach to our adversarial configuration, however they utilize the Wasserstein GAN objective \cite{pmlr-v70-arjovsky17a} rather than the basic GAN formulation that we use.  The authors additionally use the same approximation that we utilize in section \ref{sec:discontinuity}. We note that they employ other strategies such as teacher helping, curriculum learning, and variable length that are worth looking into for future work. We note, however, that the authors tackle general language generation, rather than our task of \hyperref[def:contextualcommonsenseinference]{contextual commonsense inference}.

\section{Hinting for Controllable Generation}
In this section, we propose a technique called “Hinting” to address the lack of controllability in \hyperref[def:contextualcommonsenseinference]{contextual commonsense inference}.

\subsection{What is hinting?}
\label{sec:whatishinting}

Recently, there has been work on exploring \textit{prompting} strategies \cite{liu2021pre} for pre-trained, transformer-based language models \cite{vaswani2017attention,devlin-etal-2019-bert}.  These are methods which alter the input to a language model such that it matches or approximates templates that it has seen during pre-training and can reuse or exploit this information.  Prompting helps achieve higher performance in tasks with less training data,  can help with controllability in the case of text generation, and is more parameter-efficient and data-efficient than fine-tuning, in some cases \cite{li2021prefix}. One type of prompting is \textit{prefix prompting} \cite{li2021prefix,lester2021power}.  Prefix prompting consists of altering a language model's input (i.e. prefix) by adding additional words.  These words can be explicit hard prompts such as actual phrases or words, or they can be soft prompts, embeddings that are input into a model and can be trained to converge on some virtual template or virtual prompt that can help the model.  

Prompting holds potential for improving \hyperref[def:contextualcommonsenseinference]{contextualized commonsense inference}.  We utilize the idea of a \textit{hint}, a hybrid of hard and soft prompts. We define a \hyperref[def:hint]{\textit{hint}} as the part(s) of an assertion that a
model has to predict, along with special identifiers for these parts,
wrapped within parenthesis characters. A forthcoming companion paper focuses on the \hyperref[def:hinting]{hinting} mechanism. We include a brief description of it here, but we note that the focus of this work is the adversarial generation with the \hyperref[def:jointcommonsenseinference]{joint inference} training. Because \hyperref[def:hinting]{hinting} is an essential component for controllability of this model, we include this overview. 

Syntactically, a \hyperref[def:hint]{\textit{hint}}
 takes the form of: “([specificity], [subject symbol,subject],
[relation symbol,relation], [object symbol,object] )” where the actual
content of the hint between the parenthesis would be a
permutation of all but one of the elements in the target tuple.
 Hints are provided during training by sampling a binomial distribution ($p=0.5$) for each element in a minibatch, which
determines whether to give a \hyperref[def:hint]{hint} or not.  The actual content
of the \hyperref[def:hint]{hint} would then be generated by randomly sampling without replacement up to all but one of the elements in a target tuple. In Table \ref{tab:illustrativeexample} we can see various examples of how the \hyperref[def:hinting]{hinting} mechanism works for a model trained on GLUCOSE.

\begin{table*}[h]
\resizebox{\textwidth}{!}{

\begin{tabular}{|l|l|}
\hline
Model Input                                                                                                                                                                                                                                                                                                                                                      & Model Output                                                                                                                                                                                                                                                                     \\ \hline
\begin{tabular}[c]{@{}l@{}}7: The hockey game was tied up.  \\ The red team had the puck. They sprinted down the ice. \\ They cracked a shot on goal!. * They scored a final goal!. *\end{tabular}                                                                                                                        & \begin{tabular}[c]{@{}l@{}}They scored a final goal \textgreater{}Causes\textgreater They feel(s) happy ** \\ Some People\_A scored a final goal \textgreater{}Causes\textgreater Some People\_A feel(s) happy\end{tabular}                                                      \\ \hline
\begin{tabular}[c]{@{}l@{}}7: The hockey game was tied up.  \\ The red team had the puck. They sprinted down the ice. \\ They cracked a shot on goal!. * They scored a final goal!. *\\ \textbf{hint: (\textless{}|specific|\textgreater{}\textless{}|subj|\textgreater \textcolor{Emerald}{the red team scores the final goal})}\end{tabular}          & \begin{tabular}[c]{@{}l@{}}\textcolor{Emerald}{\textbf{the red team scores the final goal}} \textgreater{}Causes\textgreater the red team feel(s) happy ** \\ Some People\_A (who are a team) score the final goal \textgreater{}Causes\textgreater Some People\_A feel(s) happy\end{tabular}                  \\ \hline
\begin{tabular}[c]{@{}l@{}}7: The hockey game was tied up.  \\ The red team had the puck. They sprinted down the ice.\\  They cracked a shot on goal!. * They scored a final goal!. *\\ \textbf{hint: (\textless{}|specific|\textgreater{}\textless{}|subj|\textgreater \textcolor{Emerald}{the blue team does not score the final goal})}\end{tabular} & \begin{tabular}[c]{@{}l@{}}\textcolor{Emerald}{\textbf{the blue team does not score the final goal}} \textgreater{}Causes\textgreater the blue team feel(s) disappoinment ** \\ Some People\_A do not score the final goal \textgreater{}Causes\textgreater Some People\_A feel(s) disappointment\end{tabular} \\ \hline
\begin{tabular}[c]{@{}l@{}}7: The hockey game was tied up.  \\ The red team had the puck. They sprinted down the ice. \\ They cracked a shot on goal!. * They scored a final goal!. *\\ \textbf{hint: (\textless{}|specific|\textgreater{}\textless{}|obj|\textgreater \textcolor{Peach}{a child})}\end{tabular}                                      & \begin{tabular}[c]{@{}l@{}}they scored a final goal \textgreater{}Causes\textgreater \textcolor{Peach}{\textbf{a child} feel(s) happy} ** \\ Some People\_A scored a final goal \textgreater{}Causes\textgreater Someone\_A feel(s) happy\end{tabular}                                                       \\ \hline
\begin{tabular}[c]{@{}l@{}}7: The hockey game was tied up.  \\ The red team had the puck. They sprinted down the ice. \\ They cracked a shot on goal!. * They scored a final goal!. *\\ \textbf{hint: (\textless{}|general|\textgreater{}\textless{}|subj|\textgreater\textcolor{Emerald}{\textbf{ Something\_A (that is a point)}})}\end{tabular}               & \begin{tabular}[c]{@{}l@{}}They scored a final goal \textgreater{}Causes\textgreater They feel(s) happy ** \\ \textcolor{Emerald}{\textbf{Something\_A (that is a point))} is scored} \textgreater{}Causes\textgreater Some People\_A feel(s) happy\end{tabular}                                               \\ \hline
\end{tabular}}
\caption{Example of inputs and outputs for the GLUCOSE trained contextual commonsense inference model with hints. The \textit{hint} is \textbf{bolded} and the parts of the \textit{hint} are colored (\textcolor{Emerald}{subject}, \textcolor{Orchid}{relation}, \textcolor{Peach}{object}). Without a hint we can see that the model tries to infer directly on the content of the sentence, however with \textit{hints}, the model tries to include an inference based on the target sentence with the contents of the \textit{hint}.}
\label{tab:illustrativeexample}
\end{table*}

Here we can see more clearly that whenever we give a \hyperref[def:hint]{hint} a model trained with \hyperref[def:hint]{hints} (i.e., \hyperref[def:hinting]{hinting}) tends to produce generations that include the components given in the \hyperref[def:hint]{hint}.  We utilize \hyperref[def:hinting]{hinting} in training our models from here on out unless otherwise stated.  The controllability that \hyperref[def:hinting]{hinting} enables can permit us to use models trained with it in downstream applications such as contextual \hyperref[def:knowledgegraph]{knowledge graph }generation.

\subsubsection{Discourse-aware/contextual commonsense inference}
\label{sec:contextualcommonsenseinference}
Recall that commonsense inference is the task of generating a commonsense assertion.  \hyperref[def:contextualcommonsenseinference]{Discourse-aware/contextual commonsense inference} is the task of, given a certain context, inferring commonsense \hyperref[def:assertion]{assertions} that are coherent within the narrative \cite{Gabriel_PC}. This task is particularly hard because commonsense knowledge may not be explicitly stated in text \cite{liu2004conceptnet} and the model needs to keep track of entities and their states either explicitly or implicitly.  Research into the knowledge that pre-trained language models learn has yielded good results in that they do contain various types of factual knowledge, as well as some commonsense knowledge \cite{da2019cracking,petroni2019language,davison2019commonsense}. The amount of commonsense knowledge in these models can be improved by supplementing sparsely covered subject areas with structured knowledge sources such as ConceptNet \cite{speer2017conceptnet,davison2019commonsense}. Knowing that these pre-trained language models may contain some commonsense information has led to the development of knowledge models such as COMET \cite{bosselut2019comet}.  This line of research has been extended from the  sentence-by-sentence level in COMET, to the paragraph-level in ParaCOMET \cite{Gabriel_PC}. Contemporaneously, GLUCOSE \cite{mostafazadeh-etal-2020-GLUCOSE}   builds a dataset of commonsense \hyperref[def:assertion]{assertions} that are contextualized to a set of stories, and \hyperref[def:generalinference]{generalized} (e.g., \textit{John is a human} is generalized to \textit{Someone\_A is a human}). 

The general task of \hyperref[def:contextualcommonsenseinference]{contextual commonsense inference} can be formally described as follows.  We are given a story $S$ composed of $n$ sentences, $S=\{S_1,S_2,…,S_n\}$ , a target sentence from that story, $S_t$, where $S_t \in S$, and a relation type $R$. Given all this, we want to generate a tuple in the form of $(specificity, \textcolor{Emerald}{subject}, \textcolor{Orchid}{R} , \textcolor{Peach}{object})$ that represents an assertion, present or implied, in $S_t$ given the context $S$, and the relation type $R$. 

\subsubsection{An example of Hinting}
\label{sec:hinting}
A simple example of \textit{hinting} is the following:
\begin{quote}
    \small
    \textbf{Story:} \textit{The hockey game was tied up. The red team had the puck. They sprinted down the ice. They cracked a shot on goal! They scored a final goal!} 
    
    \textbf{Target sentence:} \textit{They scored a final goal!}

    \textbf{Target assertion:} \textit{(subject: \textcolor{Emerald}{the red team}, relation: \textcolor{Orchid}{are capable of}, object: \textcolor{Peach}{winning the game}.)}
\end{quote}

A \hyperref[def:hint]{hint} can be any permutation of the target \hyperref[def:assertion]{assertion}, except the complete \hyperref[def:assertion]{assertion}, along with some symbol that indicates which part it is:
\begin{quote}
    \small
\textbf{Possible Hints:}\textit{ (<|subj|> \textcolor{Emerald}{the red team}), (<|subj|> \textcolor{Emerald}{the red team},  <|rel|> \textcolor{Orchid}{capable of}), (<|subj|> \textcolor{Emerald}{the red team}, <|obj|> \textcolor{Peach}{winning the game}), (<|rel|> \textcolor{Orchid}{capable of}, <|obj|> \textcolor{Peach}{winning the game}), (<|obj|> \textcolor{Peach}{winning the game}), (<|rel|> \textcolor{Orchid}{capable of})}
\end{quote}

A \hyperref[def:hint]{hint} for the given story, target sentence and target \hyperref[def:assertion]{assertion}, yields the following:
\begin{quote}
    \small
    \textbf{Hint:} \textit{(<|subj|> \textcolor{Emerald}{the red team},  <|rel|> \textcolor{Orchid}{capable of})}

\end{quote}

Putting everything altogether, the input for the model would be: 

\begin{quote}
    \small
    \textbf{Story with Hint:} \textit{The hockey game was tied up. The red team had the puck. They sprinted down the ice. They cracked a shot on goal! They scored a final goal! \textbf{(<|subj|> \textcolor{Emerald}{the red team},  <|rel|> \textcolor{Orchid}{capable of})}}. 
\end{quote}

We note that this is a generic version of how the \hyperref[def:hinting]{hinting} mechanism works, and individual datasets (i.e., ParaCOMET and GLUCOSE) have slightly different variations of this. 

\subsection{Experimental Setup}
We run two sets of experiments to show the effectiveness of \hyperref[def:hinting]{hinting}.  The first is utilizing the original ParaCOMET dataset and setup and adding \hyperref[def:hint]{hints}. The ParaCOMET setup consists of given a story $S$ composed of $n$ sentences, $S=\{S_1,S_2,…,S_n\}$, a relation type $R$, and a target sentence token (i.e. $<|sent0|>$, $<|sent1|>$, …, $<|sent(n-1)|>$). In the ParaCOMET dataset, we must predict the \textcolor{Peach}{object} of a triple, utilizing implicitly the sentence as a  \textcolor{Emerald}{subject} and explicitly the supplied sentence symbol and  \textcolor{Orchid}{relation} $R$ symbol. 

Within this framework, after the relation $R$, we add our \hyperref[def:hint]{hint} between parenthesis (i.e. “([\textit{hint}])”).  In this framing, our \hyperref[def:hint]{hint} can be composed of: a subject symbol (<|subj|>) along with the target sentence to serve as a \textcolor{Emerald}{subject}, a relation symbol along with the \textcolor{Orchid}{relation}  $R$, or an object symbol along with the \textcolor{Peach}{object} of the triple. Using the hockey example, a possible \hyperref[def:hint]{hint} in this set of experiments would be: “(\textcolor{Orchid}{<|rel|> <|xEffect|>},\textcolor{Peach}{<|obj|> they win the game})”.  

In our experiments on the ParaCOMET formulation with the GPT-2 model, we utilize the same cross-entropy loss as in \cite{Gabriel_PC}.  We note that we utilize a sequence-to-sequence \cite{sutskever2014sequence} formulation for the T5 and the BART models. This in contrast to the GPT-2-based system requires encoding a source sequence (i.e., story, target sentence, and relation symbol), and decoding it into a target sequence (i.e., the \textcolor{Peach}{object} of an assertion). For the T5 model, we add the prefix “source:” before the story $S$, and the prefix “hint:” for placing our \hyperref[def:hint]{hints}. For simplicity, we construct the same “heuristic” dataset as ParaCOMET  which utilizes a heuristic matching technique to align ATOMIC \cite{sap2019ATOMIC} triples to story sentences. 

For our second set of experiments, we utilize the formulation utilized in GLUCOSE \cite{mostafazadeh-etal-2020-GLUCOSE}.  The formulation utilizes the T5 model in a sequence-to-sequence formulation once more.  In this formulation, the source text is composed of a prefix of a dimension to predict $D \in {1,2,…10}$\footnote{The definition for each dimension number is given in the GLUCOSE work. Dimensions in GLUCOSE are (explicit or implicit) relations that help explain causality between the entities mentioned.}, followed by the story $S$ with the marked target sentence. The target sentence, $S_t$, is marked with $*$ before and after the sentence. An example input is: "1: The first sentence. *The target sentence. * The third sentence.". This task is slightly different from the ParaCOMET one, in that in addition to predicting a \hyperref[def:storyspecificinference]{context specific assertion}, the model has to predict a \hyperref[def:generalinference]{generalized assertion} (i.e., in this task we have to infer a general and context specific  \textcolor{Emerald}{subject}, \textcolor{Peach} {object} and a  \textcolor{Orchid}{relation}). For our \hyperref[def:hint]{hints} we provide up to five out of these six things, along with a symbol that represents whether it is the \textcolor{Emerald}{subject}, \textcolor{Peach} {object} or a  \textcolor{Orchid}{relation}, and another symbol that represents whether it is part of the general or specific assertion. We add our \textit{hint} after the story $S$, utilizing the prefix “hint:” and supplying the \hyperref[def:hint]{hints} between parenthesis. 

We run the ParaCOMET experiments for 10 epochs on the dataset's training data and evaluation data. We utilize a max source sequence length for the BART and T5 models of 256, and a max target length of 128. For the GPT-2 models we utilize a max sequence length of 384.  
Additionally, we use the ADAM \cite{kingma2014adam} optimizer with a learning rate of 2e-5, and a linear warm-up of 0.2 percent of the total iterations. For the T5 models we utilize a learning rate of 1e-4 because early experiments showed that the model would not converge with lesser learning rates.  We utilize the scripts from \cite{Gabriel_PC} for data generation. We also utilize a batch size of 4 for training and we accumulate gradients for 4 steps for an effective batch size of 16. The results that we present are the average of the 10 runs over 4 seeds for hinted and non-hinted conditions.

We run GLUCOSE experiments for 5 epochs and 4 seeds on the original  GLUCOSE data. Additionally, we utilize a linear warm-up of 3000 steps. We utilize the ADAM optimizer with a learning rate of 3e-4, a train batch size 4, with gradient accumulation of 4 steps for an effective batch size of 16, and a max source length of 256 and max target length of 128. In our results, we present the average of the 4 seeds across the 5 epochs. In both experiments we report the scores given by  SacreBLEU \cite{post-2018-call}, ROUGE \cite{lin-2004-rouge}, and METEOR \cite{banarjee2005} using the datasets library \cite{quentin_lhoest_2021_5570305} metrics system. We run our experiments in a machine with an AMD ThreadRipper 3970 Pro and 4 NVIDIA A6000s. Every epoch per model is approximately an hour.

Additionally, we run a small Mechanical Turk study similar to the one presented in the original ParaCOMET \cite{Gabriel_PC} in which a human judges a generated assertion and judges the plausibility of it on a  5-point Likert scale: obviously true (5), generally true (4), plausible (3), neutral or unclear (2), and doesn’t make sense (1). We present the results in the same manner where Table \ref{tab:expset3} displays the percent of inferences judged as plausible or true (3-5), and the average rating per inference. Participants were given \$0.1  to complete the task. We sample from each of the ParaCOMET and GLUCOSE test sets, 100 entries. Then based on the models for each dataset, we pick the epoch that had the highest automated scores and we proceed to randomly sample one of the trained hint and non-hinted models. We then select one sentence of the randomly sampled test entries and ask both models to generate an inference along a randomly sampled relation or dimension for that sentence.  
\subsection{Results and Effects of hinting}

\begin{table*}[h]
\resizebox{\textwidth}{!}{
\begin{tabular}{|l|ll|ll|ll|ll|ll|ll|}
\hline
Model              & \multicolumn{2}{l|}{BLEU}                               & \multicolumn{2}{l|}{METEOR}                             & \multicolumn{2}{l|}{ROUGE1}                             & \multicolumn{2}{l|}{ROUGE2}                             & \multicolumn{2}{l|}{ROUGE L}                            & \multicolumn{2}{l|}{ROUGE L SUM}                        \\ \hline
                   & \multicolumn{1}{l|}{Hint}             & No Hint         & \multicolumn{1}{l|}{Hint}             & No Hint         & \multicolumn{1}{l|}{Hint}             & No Hint         & \multicolumn{1}{l|}{Hint}             & No Hint         & \multicolumn{1}{l|}{Hint}             & No Hint         & \multicolumn{1}{l|}{Hint}             & No Hint         \\ \hline
\textbf{ParaCOMET} & \multicolumn{1}{l|}{\textbf{42.705*}} & 41.960 & \multicolumn{1}{l|}{\textbf{59.411*}} & 59.045 & \multicolumn{1}{l|}{\textbf{63.339*}} & 61.454 & \multicolumn{1}{l|}{\textbf{52.483*}} & 50.513          & \multicolumn{1}{l|}{\textbf{63.292*}} & 61.395          & \multicolumn{1}{l|}{\textbf{63.294*}} & 61.399          \\ \hline
\textbf{Bart}      & \multicolumn{1}{l|}{\textbf{41.765*}} & 41.639 & \multicolumn{1}{l|}{\textbf{58.766*}} & 58.639 & \multicolumn{1}{l|}{\textbf{61.054}}  & 61.013 & \multicolumn{1}{l|}{\textbf{49.970}}  & 49.889          & \multicolumn{1}{l|}{\textbf{61.004}}  & 60.964          & \multicolumn{1}{l|}{\textbf{61.010}}  & 60.969          \\ \hline
\textbf{T5}        & \multicolumn{1}{l|}{41.070}           & \textbf{41.102} & \multicolumn{1}{l|}{\textbf{58.004}}  & 58.000          & \multicolumn{1}{l|}{59.535}           & \textbf{59.631} & \multicolumn{1}{l|}{48.695}           & \textbf{48.823} & \multicolumn{1}{l|}{59.488}           & \textbf{59.588} & \multicolumn{1}{l|}{59.494}           & \textbf{59.597} \\ \hline
\end{tabular}}
\caption{Averages of 4 different seeds over 10 epochs for \textit{hinted} (Hint) and non-hinted (No Hint) runs of the ParaCOMET dataset from \cite{Gabriel_PC}.  The largest scores are \textbf{bolded} and significantly different scores have an asterisk (*) next to them. 
% We evaluated the task on the original ParaCOMET (GPT-2) model, and on a Bart and T5 model to evaluate the effect of hinting on a sequence-to-sequence encoder-decoder model.
We can see from the results that \textit{hinted} systems tend to achieve higher performance even if slightly and in some cases significantly, and do not decrease performance significantly. For significance we use the t-Test: Paired Two Sample for Means.}
\label{tab:expset1}
\end{table*}

% Please add the following required packages to your document preamble:
% 
\begin{table*}[h]
\resizebox{\textwidth}{!}{\begin{tabular}{|ll|ll|ll|ll|ll|ll|}
\hline
\multicolumn{2}{|l|}{BLEU}                       & \multicolumn{2}{l|}{Meteor}                    & \multicolumn{2}{l|}{Rouge 1}                    & \multicolumn{2}{l|}{Rouge 2}                   & \multicolumn{2}{l|}{ROUGE L}                    & \multicolumn{2}{l|}{Rouge LSUM}                 \\ \hline
\multicolumn{1}{|l|}{No Hint} & Hint             & \multicolumn{1}{l|}{No Hint} & Hint            & \multicolumn{1}{l|}{No Hint} & Hint             & \multicolumn{1}{l|}{No Hint} & Hint            & \multicolumn{1}{l|}{No Hint} & Hint             & \multicolumn{1}{l|}{No Hint} & Hint             \\ \hline
\multicolumn{1}{|l|}{58.542}  & \textbf{59.099*} & \multicolumn{1}{l|}{66.829}  & \textbf{66.917} & \multicolumn{1}{l|}{66.387}  & \textbf{66.681*} & \multicolumn{1}{l|}{47.850}  & \textbf{48.141} & \multicolumn{1}{l|}{62.542}  & \textbf{62.874*} & \multicolumn{1}{l|}{62.528}  & \textbf{62.868*} \\ \hline
\end{tabular}}

\caption{Averages of 4 different seeds over 5 epochs for \textit{hinted} (Hint) and non-hinted (No Hint) runs of the GLUCOSE contextual inference task dataset. This is the same dataset as the work in \cite{mostafazadeh-etal-2020-GLUCOSE}  The largest scores are \textbf{bolded} and significantly different scores have an asterisk (*) next to them. Once more, we see that hinting provides a small, increase in performance, all the while permitting controllability. t-Test: Paired Two Sample for Means}
\label{tab:expset2}
\end{table*}

\begin{table*}[h]
    \begin{minipage}{.5\linewidth}
      \centering
       \small \begin{tabular}{|l|l|l|}
\hline
Model      & Non-Hinted & Hinted \\ \hline
\textbf{ParaCOMET}  & 3.71       & \textbf{3.76}   \\ \hline

\textbf{Bart}       & \textbf{3.72}       & 3.48   \\ \hline
\textbf{T5}         & \textbf{3.73}       & 3.68   \\ \hline
\textbf{T5-GLUCOSE} & \textbf{4.10}       & 4.06   \\ \hline
\end{tabular}
    \end{minipage}%
    \begin{minipage}{.5\linewidth}
      \centering
       \small \begin{tabular}{|l|l|l|}
\hline
Model      & Non-Hinted & Hinted \\ \hline
\textbf{ParaCOMET}  & 81\%         & \textbf{84\%}     \\ \hline

\textbf{Bart}       & \textbf{83\%}         & 74\%     \\ \hline
\textbf{T5}         & \textbf{81\%}         & \textbf{81\%}     \\ \hline
\textbf{T5-GLUCOSE} & \textbf{92\%}         & 90\%     \\ \hline
\end{tabular}
    \end{minipage} 
        \caption{Results of human evaluation of ParaCOMET and GLUCOSE datasets. The largest scores are \textbf{bolded} and significantly different scores have an asterisk (*) next to them. We sampled 100 test points for each model from their test datasets and had the hinted and non-hinted models infer assertions. Humans judged these assertions on a 5 point Likert scale where above 3 was plausible similar to  \cite{Gabriel_PC}. On the left we can see the average values of the human judgments and on the right we can see the percentage of plausible inferences (rated >= 3). We can see that hinting provides comparable performance.}
\label{tab:expset3}

\end{table*}

\subsubsection{Experiment 1: ParaCOMET with \textit{hints}}
The aggregated (averaged) results for this set of experiments can be found in Table \ref{tab:expset1}. We can see here that on average, \hyperref[def:hinting]{hinting} does tend to improve the score even if slightly.  It seems that providing a \hyperref[def:hint]{hint} is beneficial and not detrimental for contextual commonsense inference.   Given the way that this task is framed, a possibility that could explain the relative similarity of the performances, is that \hyperref[def:hinting]{hinting} \textit{only} adds the \textcolor{Peach}{object} of the \hyperref[def:assertion]{assertion} as additional possible data that the model may see during training; the  \textcolor{Emerald}{subject} and the  \textcolor{Peach}{relation} can be repeated with \hyperref[def:hinting]{hinting}.   We note that the performance of the T5 model was less than that of the other models, and we believe that it may be lack of hyperparameter tuning, as it was seen that the model was sensitive to the learning rate and had to use a higher than usual learning rate.  

\subsubsection{Experiment 2: GLUCOSE with \textit{hints}}
The aggregated results for this set of experiments can be found in Table \ref{tab:expset2}. Once more, we notice that \hyperref[def:hinting]{hinting} does tend to improve the performance of the \hyperref[def:contextualcommonsenseinference]{contextual commonsense inference} task.  This suggests that \hyperref[def:hinting]{hinting} is indeed beneficial for the task, especially when faced with the harder task of generating both a \hyperref[def:generalinference]{general} and \hyperref[def:storyspecificinference]{specific} \hyperref[def:assertion]{assertion}.  We believe that this improvement is because \hyperref[def:hinting]{hinting} gives the model the clues it may need to decide on what to focus or attend to, to generate useful inferences, but further experimentation would be needed to verify this. 

\subsubsection{Experiment 3: Human Judgements}
The results for a small Mechanical Turk study for human evaluation of model inferences can be seen in Table \ref{tab:expset3}.  Overall, we can see here that \hyperref[def:hint]{hinted} systems are judged as less plausible. Interestingly, after inspecting the results where there was a large difference (more than two points between the systems), we see that there are some cases in which the same or very similar responses got completely different scores. We also see upon looking some of the inferences that the hinted model tends to be more general and provide shorter  responses than the non-hinted model (e.g., hinted inference: “satisfied” vs. non-hinted inference: “happy and satisfied”). 
% This could possibly be that the model learns to be specific when hinted, and general when not, however more testing would be needed to confirm this.

\subsubsection{Final Remarks on Hinting}
From the results of our experiments, we can see that \hyperref[def:hinting]{hinting} tends to increase the performance of \hyperref[def:contextualcommonsenseinference]{contextualized commonsense inference} at least with regard to automated metrics and does not significantly degrade or improve human judgements.  Without any significant cost, by utilizing hinting, we gain controllability in the generation. By supplying these \textit{hints}, we are teaching the model to pay attention and generate inferences about a certain \textcolor{Emerald}{subject}, \textcolor{Orchid}{relation}, or  \textcolor{Peach}{object}. This in turn, after training, can be leveraged by a user or downstream application to guide the model to generate \hyperref[def:assertion]{assertions} from parts that are manually supplied. Although this is not very clear within the ParaCOMET formulation, it becomes clearer in the GLUCOSE formulation of the problem.  We give an illustrative example of the usefulness of \hyperref[def:hinting]{hinting} in Table \ref{tab:illustrativeexample}. We can see that by giving a model the \hyperref[def:hint]{hint}, the model could be capable of inferring about information that may not be present in the story. We note that this behavior is useful in downstream tasks such as story understanding and contextual \hyperref[def:knowledgegraph]{knowledge graph }generation, in which we may need a model to have a specific \textcolor{Emerald}{subject} or \textcolor{Peach}{object}.

\section{Joint Inference of Commonsense Assertions}
\label{sec:jointhinting}
Given that we have presented \hyperref[def:hinting]{hinting}, which is a method to help with the controllability of \hyperref[def:contextualcommonsenseinference]{contextual commonsense inference} generation, we now look at how we can combine multiple knowledge bases for this task along with \hyperref[def:hinting]{hinting} for \hyperref[def:contextualcommonsenseinference]{contextual commonsense inference}. 
\subsection{What is joint commonsense inference?}
In this work, we define \hyperref[def:jointcommonsenseinference]{\textit{Joint commonsense inference}} as inferring commonsense knowledge \hyperref[def:assertion]{assertions} by leveraging knowledge from multiple knowledge bases.   To illustrate this, we give the following example, story: 
\begin{quote}
\centering
   \textit{John is a regular person who has a dog. \textbf{John, every day, goes out to walk his dog.} John met a friend when walking his dog. They exchanged stories about their dogs.}
\end{quote}
From this story, we want to infer the \hyperref[def:generalinference]{general} version of the commonsense assertion of “John is capable of walking his dog”, derived from \textbf{the second sentence}.  This \hyperref[def:generalinference]{general} version can look similar to “Someone\_A who has an animal (that is a dog) enables Someone\_A to walk the animal (that is a dog)”.  To generalize this, we must know that: \textit{John is a person's name}, which we can find from a semantic tagger. A much more commonsensical fact needed to infer the \hyperref[def:assertion]{assertion} is that: \textit{a dog is an animal}, which is a fact found in ConceptNet. Lastly, to infer the \hyperref[def:assertion]{assertion}, we need to know that: \textit{A person having a dog has the effect that a person goes to walk their dog}, which is a fact that we could find from ATOMIC 2020. Therefore, to infer the general assertion that we presented, we must join information from \textit{at least} two knowledge bases to infer our \hyperref[def:generalinference]{general assertion}.  This process of joining the information from multiple sources is what we call \hyperref[def:jointcommonsenseinference]{joint commonsense inference}. \hyperref[def:jointcommonsenseinference]{Joint commonsense inference} is useful because it could lead to implicitly applying or combining knowledge and/or analogies that might be present in the different knowledge sources, which may lead to better results when performing \hyperref[def:contextualcommonsenseinference]{contextual commonsense inference}.

\subsection{Joint Inference Approach Overview}
To perform \hyperref[def:jointcommonsenseinference]{joint inference} in the task of \hyperref[def:contextualcommonsenseinference]{contextual commonsense inference}, we propose the following approach:
\begin{quote}
    \begin{enumerate}
    \item \textit{For each knowledge base that we have, we convert each of the \hyperref[def:assertion]{assertion} found in them into a tuple format of \textbf{\{subject, relation, object, specificity\}}\footnote{This follows a similar pattern to subject-verb-object triples, but has the added field of specificity which is whether the assertion is contextual to a story, or a generally applicable assertion}. We note that each part of the tuple must be text\footnote{This ultimately helps us express the assertion in a textual way (i.e., (a dog, IsA, animal) when converted to the tuple (a dog, is a, animal, specific) and passed to a string representation function can be expressed as “Specifically, a dog is a animal”.} (i.e., if a relation is symbolically “IsA”, the text version would be “is a”).}
    \item \textit{We align each knowledge base tuple with a story (e.g., the ROCStories corpus) and target sentence from the story.  The target sentence is the sentence which is most likely to be used to infer the tuple. We perform the alignment by vectorizing the tuples and stories and utilizing nearest neighbors with the cosine distance as a metric. We give details of this alignment in section \ref{sec:aligningjointinference}.}
    \item \textit{We combine into one list and shuffle the aligned knowledge base tuples from multiple knowledge sources.}
    \item \textit{We replace the naming scheme of variables that may be present in \textit{general} \hyperref[def:assertion]{assertions} with the naming scheme from GLUCOSE (e.g., \textit{PersonX} is replaced with \textit{Person\_A}).}
    \item \textit{We train a \hyperref[def:contextualcommonsenseinference]{contextual commonsense inference} model on this dataset, whose inferences are joint inferences.}
\end{enumerate}

\end{quote}
By following this procedure, we will end up with a dataset of story aligned \hyperref[def:assertion]{assertions}.  In this dataset, all of the \hyperref[def:assertion]{assertions} are grounded in the same set of stories. With this we can train models that can perform joint \hyperref[def:contextualcommonsenseinference]{contextual commonsense inference}. Now we will go into some details of this process.  
\subsection{Specificity in \hyperref[def:assertion]{assertions}}
Recall that we define \hyperref[def:specificity]{\textit{specificity}} as whether an assertion's content is about \hyperref[def:storyspecificinference]{specific} entities in the aligned story, or if it is a \hyperref[def:generalinference]{generalized version} of an \hyperref[def:assertion]{assertion}. This can be seen as whether the assertion is a \textit{general} template with variables, or a \textit{specific} instance of this template. To make the difference between \textit{specific} and \textit{general} \hyperref[def:assertion]{assertions} clearer, we give the following example. Using the same story as before:  
\begin{quote}
   \textit{John is a regular person who has a dog. \textbf{John, every day, goes out to walk his dog.} John met a friend when walking his dog. They exchanged stories about their dogs.}
\end{quote}
As before, we focus on the second sentence: \textbf{John, every day, goes out to walk his dog.}. From here, we can infer the \textit{specific} assertion: “\textcolor{Brown}{John} is capable of walking his \textcolor{magenta}{dog}”. The assertion is \textit{specific} because it fills out a broadly applicable template, which we will present next, that speaks about John and his dog from the story.  From the sentence, we can also infer the \textit{general} version of the assertion: “\textcolor{Brown}{Someone\_A} who has \textcolor{magenta}{Something\_A (that is a dog)} enables \textcolor{Brown}{Someone\_A} to walk the \textcolor{magenta}{Something\_A (that is a dog)}”. This latter assertion is \textit{general} because it speaks in a template format (i.e., broader terms) the same fact.  A \textit{general} assertion is not the story-dependent instance of the template, but the broader, story-independent template.  These \textit{general} \hyperref[def:assertion]{assertions} contain variables in them. 

In this work we utilize ConceptNet, ATOMIC 2020, and GLUCOSE as our knowledge bases, and propose to combine them to perform \hyperref[def:jointcommonsenseinference]{joint inference}.  In table \ref{tab:specificity} we give the different available \hyperref[def:specificity]{specificities} for these knowledge bases. 
\begin{table}[h]
\begin{tabular}{|l|l|l|}
\hline
\textbf{Knowledge Base} & \textbf{General} & \textbf{Specific} \\ \hline
ConceptNet              & \textcolor{red}{\xmark}*                & \textcolor{green}{\cmark}                 \\ \hline
ATOMIC 2020             & \textcolor{green}{\cmark}                & \textcolor{red}{\xmark}*                 \\ \hline
GLUCOSE                 & \textcolor{green}{\cmark}                & \textcolor{green}{\cmark}                 \\ \hline
\end{tabular}
\caption{Here we can see the available specificities in ConceptNet, ATOMIC 2020, and GLUCOSE.  We mark with \textcolor{red}{\xmark} the specificities that are not available by default, and add * to those that can be generated.}
\label{tab:specificity}
\end{table}
From this, we can see that ConceptNet does not have \textit{general} specificity \hyperref[def:assertion]{assertions}. Although this may sound counterintuitive, ConceptNet gives \textit{specific}, untemplated, instances of \hyperref[def:assertion]{assertions}, in contrast to ATOMIC and GLUCOSE, which describe \textit{general} versions of \hyperref[def:assertion]{assertions}. ATOMIC 2020 has the opposite problem, it gives  \textit{general} versions of rules, (e.g., PersonX participates in some event, has some effect on PersonX or Y around them), and does not give, within our \hyperref[def:contextualcommonsenseinference]{contextual commonsense inference} framework, the  \textit{specific} instance of the templates (e.g., filling out PersonX, PersonY, etc.). To remedy this lack of specificity within two of our sources, we mention ways to generate examples of the missing specificity, and implement the solution for ATOMIC 2020.
\subsubsection{Generating Missing Specificity}
\paragraph{\textbf{ConceptNet}}
To generate \textit{general} \hyperref[def:assertion]{assertions} for ConceptNet, we could possibly run a classifier that would determine whether a given set of tokens is a, person, place, object, among others. With this information we could fill out, as an example, the template that GLUCOSE broadly utilize which is:  \{Category\}(\{Description\}), relation, \{Possibly Other Category\} (\{Possibly Other Description\}). From ConceptNet, we could find the relation: “\textcolor{Brown}{a dog}, IsA, \textcolor{magenta}{animal}”.  A \textit{general} version of this assertion can be “\textcolor{Brown}{Something\_A (that is a dog)}, IsA, \textcolor{magenta}{Something\_B (that is a animal)}”. Although we describe this process, we do not implement it in our work.   
\paragraph{\textbf{ATOMIC 2020}} To generate \textit{specific} \hyperref[def:assertion]{assertions} for ATOMIC 2020 we can do the following.  We can first identify variables (PersonX, PersonY, etc.) that are in the assertion.  We can then replace these variables with a mask token from a language model that was trained with the masked language modeling objective \cite{devlin-etal-2019-bert,liu2019roberta}, and use the language model to fill in the Mask token similar to a cloze (i.e., fill-in-the-blanks) task. To give the model sufficient context, we insert the assertion to the right of the nearest aligned sentence (we describe the process to get this in the next section). In the case of Person\textit{N} variable, this usually leads to the variable being replaced with a character from the story. In addition to this, ATOMIC 2020 contains blanks demarcated by underscore characters (i.e., \_\_\_\_ ), which we can once more replace with a mask token and have the model fill it out with the given context. We use this process in our work, filling in the blanks with a ROBERTA \cite{liu2019roberta} large model.  

\subsection{Aligning Assertions with Stories}
\label{sec:aligningjointinference}
\begin{figure}[h]
\begin{minipage}{0.45\linewidth}
\centering
\includegraphics[width=0.9\textwidth]{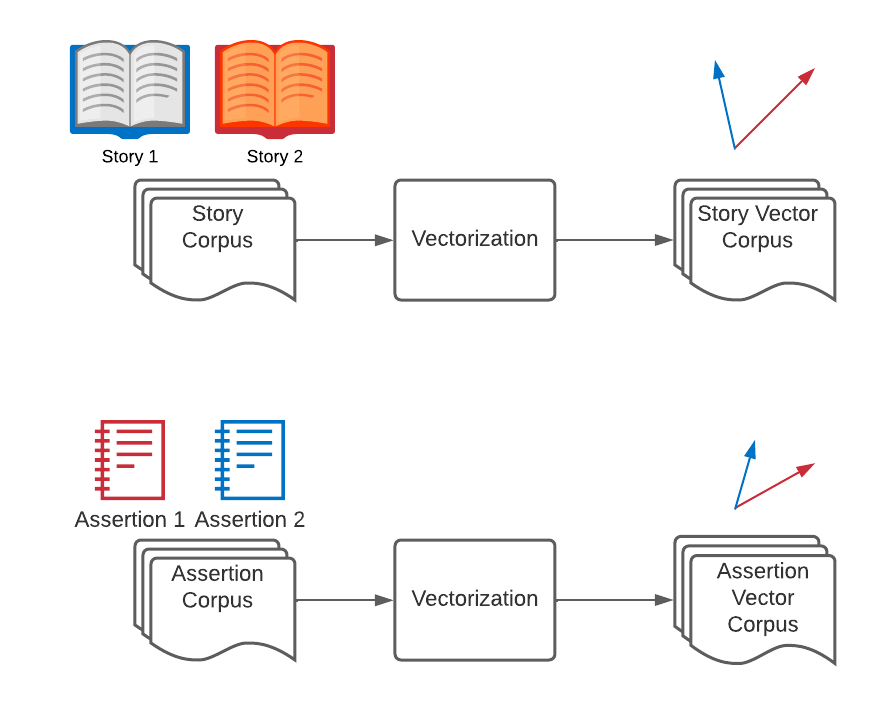}
\caption{Step 1: The story and assertion corpus are vectorized.  In our work we utilize the sentence-transformers package \cite{reimers-2019-sentence-bert} to achieve this.}\label{fig:alignment1}
\end{minipage}
\hspace{0.5cm}
\begin{minipage}{0.45\linewidth}
\centering
\includegraphics[width=\textwidth]{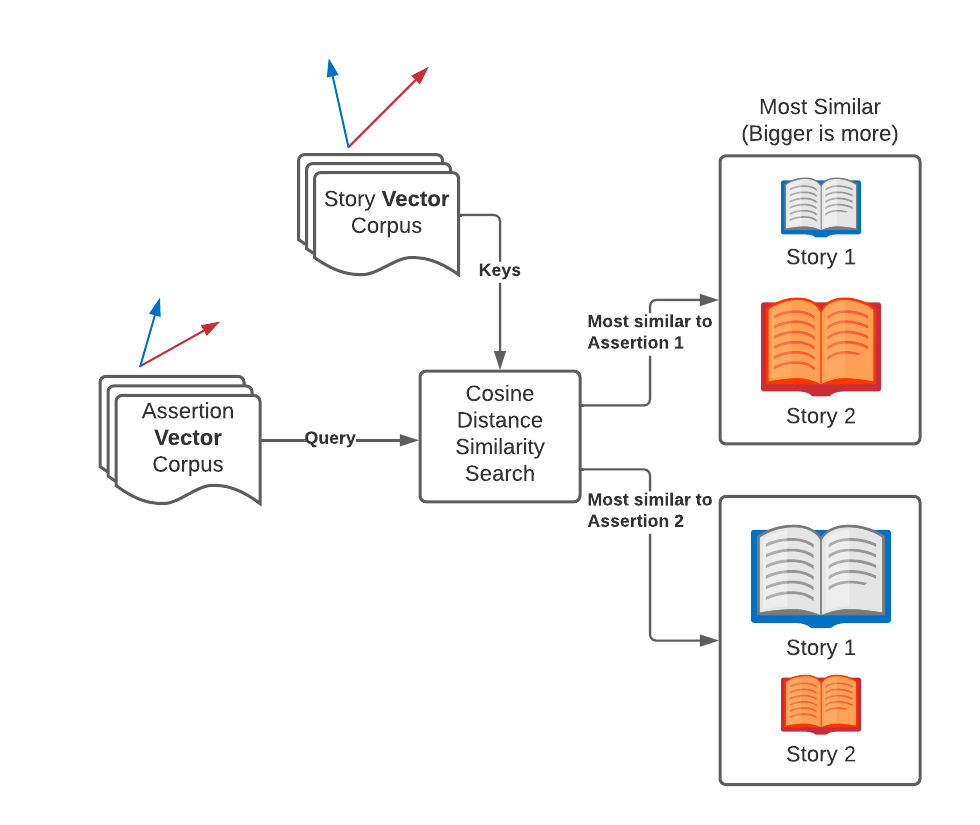}
\caption{Step 2: The resulting assertion vectors are utilized as queries, and the resulting story vectors are used as keys for a memory-like lookup. In this work we use the FAISS package for this.  The output of the memory-like lookup is the nearest story for each vector. This process is repeated for the sentences in the nearest story, to align the assertion with a sentence.}\label{fig:alignment2}
\end{minipage}
\end{figure}

To align \hyperref[def:assertion]{assertions} with stories, we do of the following procedure. On a high-level, we vectorize the stories, and we vectorize the \hyperref[def:assertion]{assertions} and we then utilize the cosine distance to find the nearest story for each \hyperref[def:assertion]{assertion}. We then go into more detail and repeat the same procedure (i.e., vectorization and similarity search) for each sentence in the previously found nearest story. Ultimately, we are left with the story and sentence that is most relevant or similar to the \hyperref[def:assertion]{assertion}.  On a low-level, we utilize the sentence transformers package along with the “paraphrase-mpnet-base-v2” model from the repository, to generate a representative vector for every story and for every \hyperref[def:assertion]{assertion} from each of our knowledge bases. We then utilize the FAISS package \cite{JDH17} to perform a fast approximate cosine similarity search to find, for each \hyperref[def:assertion]{assertion}, what is the nearest story. Once we have this nearest story, we again utilize the sentence transformers model to vectorize every sentence in that story along with the FAISS package for the cosine similarity search, to find the nearest sentence to the \hyperref[def:assertion]{assertion}.  This process can be visualized and figures \ref{fig:alignment1} and \ref{fig:alignment2}.

\subsection{Experimental Setup}
To evaluate the effects of \hyperref[def:jointcommonsenseinference]{joint inference} by combining multiple knowledge bases in the task of \hyperref[def:contextualcommonsenseinference]{contextual commonsense inference} we do the following. We generate a story aligned assertion dataset for each knowledge base individually (i.e., for ConceptNet, for ATOMIC 2020, and for GLUCOSE) as described in the previous sections. Once we have generated a dataset for each, we proceed to perform combinations of the datasets: ConceptNet-ATOMIC 2020, ConceptNet-GLUCOSE, GLUCOSE-ATOMIC, ConceptNet-ATOMIC-GLUCOSE. For the individual and the combined datasets we perform three sets of automated tests. One that includes \hyperref[def:hinting]{hinting} the \hyperref[def:specificity]{specificity}, \textcolor{Emerald}{subject}, and \textcolor{Orchid}{relation} during evaluation, one that includes \hyperref[def:hinting]{hinting} the  \textcolor{Emerald}{subject} during evaluation, and the other without these.  The rationale behind these setups is that we want to evaluate what the model infers without any guidance, and see what it infers with varying levels of guidance with multiple knowledge sources. To train our models, we use a batch size of 50, on 4x NVIDIA A6000, a learning rate of 1e-5 for an ADAM \cite{kingma2014adam} optimizer, and 3 epochs over the data.   

We note that the data for ConceptNet we utilize is the dataset given by \cite{li2016commonsense}, specifically the data in the “train\_600k.txt” which are approximately 600,000 examples of \hyperref[def:assertion]{assertions} from ConceptNet, and as a test set we utilize the “test.txt” that they provide.  For ATOMIC 2020 we utilize the training and testing data files provided by the authors \cite{hwang2020comet}. Lastly, for GLUCOSE we use the training and evaluation files also provided by the authors in the corresponding repository.

Additionally, we look into running a small Mechanical Turk evaluation of generated test \hyperref[def:assertion]{assertions}, because we suspect that automated metrics may hurt the model's evaluation when not using hinting. We sample 100 entries from the testing files of each knowledge base (ATOMIC 2020, ConceptNet, and GLUCOSE), and run these through a set of models trained firstly with only one of the test knowledge bases (i.e. a model trained only ConceptNet, a model trained in ATOMIC 2020, and a model trained in GLUCOSE) and secondly a  model trained with the combination of knowledge bases and evaluated with and without hinting. We take the generated inferences and ask 2 raters from Amazon Mechanical Turk to determine whether the assertion is acceptable, whether it is acceptable with the context that it was aligned with, and whether the gold standard assertion was acceptably aligned with the context.  We mark as acceptable the answers that both human annotators agree as valid and the others as invalid.         
\subsection{Effects of joint inference}

% Please add the following required packages to your document preamble:
% \usepackage[table,xcdraw]{xcolor}
% If you use beamer only pass "xcolor=table" option, i.e. \documentclass[xcolor=table]{beamer}
\begin{table}[b]
\begin{tabular}{|l|l|l|l|l|l|}
\hline
\rowcolor[HTML]{B0B3B2} 
\textbf{Training Set(s)}         & \textbf{Test Set} & \textbf{Hint}                  & \textbf{BLEU} & \textbf{METEOR} & \textbf{ROUGE} \\
\rowcolor[HTML]{9AFF99} 
ATOMIC 2020, ConceptNet, GLUCOSE & ATOMIC 2020       & No                             & 51.114        & 52.224          & 52.681         \\
\rowcolor[HTML]{9AFF99} 
ATOMIC 2020, ConceptNet          & ATOMIC 2020       & No                             & 51.164        & 52.151          & 52.713         \\
\rowcolor[HTML]{9AFF99} 
ATOMIC 2020                      & ATOMIC 2020       & No                             & 51.139        & 52.699          & 52.904         \\
\rowcolor[HTML]{9AFF99} 
ATOMIC 2020, ConceptNet, GLUCOSE & ATOMIC 2020       & Subject                        & 79.89         & 80.956          & 82.927         \\
\rowcolor[HTML]{9AFF99} 
ATOMIC 2020, ConceptNet          & ATOMIC 2020       & Subject                        & 80.046        & 81.144          & 83.087         \\
\rowcolor[HTML]{9AFF99} 
ATOMIC 2020                      & ATOMIC 2020       & Subject                        & 80.203        & 81.221          & 83.079         \\
\rowcolor[HTML]{9AFF99} 
ATOMIC 2020, ConceptNet, GLUCOSE & ATOMIC 2020       & Subject, Specificity, Relation & 87.031        & 88.172          & 89.606         \\
\rowcolor[HTML]{9AFF99} 
ATOMIC 2020, ConceptNet          & ATOMIC 2020       & Subject, Specificity, Relation & 87.091        & 88.242          & 89.645         \\
\rowcolor[HTML]{9AFF99} 
ATOMIC 2020                      & ATOMIC 2020       & Subject, Specificity, Relation & 87.095        & 88.226          & 89.621         \\\hline
\rowcolor[HTML]{FFFC9E} 
ATOMIC 2020, ConceptNet, GLUCOSE & ConceptNet        & No                             & 51.892        & 58.803          & 60.302         \\
\rowcolor[HTML]{FFFC9E} 
ATOMIC 2020, ConceptNet          & ConceptNet        & No                             & 56.136        & 62.114          & 63.532         \\
\rowcolor[HTML]{FFFC9E} 
ConceptNet, GLUCOSE              & ConceptNet        & No                             & 59.285        & 63.685          & 65.65          \\
\rowcolor[HTML]{FFFC9E} 
ConceptNet                       & ConceptNet        & No                             & 60.63         & 64.653          & 66.46          \\
\rowcolor[HTML]{FFFC9E} 
ATOMIC 2020, ConceptNet, GLUCOSE & ConceptNet        & Subject                        & 76.404        & 79.484          & 78.413         \\
\rowcolor[HTML]{FFFC9E} 
ATOMIC 2020, ConceptNet          & ConceptNet        & Subject                        & 76.7          & 79.614          & 78.617         \\
\rowcolor[HTML]{FFFC9E} 
ConceptNet, GLUCOSE              & ConceptNet        & Subject                        & 76.014        & 78.664          & 77.841         \\
\rowcolor[HTML]{FFFC9E} 
ConceptNet                       & ConceptNet        & Subject                        & 76.635        & 79.108          & 78.314         \\
\rowcolor[HTML]{FFFC9E} 
ATOMIC 2020, ConceptNet, GLUCOSE & ConceptNet        & Subject, Specificity, Relation & 92.695        & 94.253          & 94.171         \\
\rowcolor[HTML]{FFFC9E} 
ATOMIC 2020, ConceptNet          & ConceptNet        & Subject, Specificity, Relation & 92.892        & 94.286          & 94.378         \\
\rowcolor[HTML]{FFFC9E} 
ConceptNet, GLUCOSE              & ConceptNet        & Subject, Specificity, Relation & 92.729        & 94.071          & 94.109         \\
\rowcolor[HTML]{FFFC9E} 
ConceptNet                       & ConceptNet        & Subject, Specificity, Relation & 92.77         & 94.159          & 94.232         \\\hline
\rowcolor[HTML]{CBCEFB} 
ATOMIC 2020, ConceptNet, GLUCOSE & GLUCOSE           & No                             & 36.23         & 41.338          & 48.629         \\
\rowcolor[HTML]{CBCEFB} 
ConceptNet, GLUCOSE              & GLUCOSE           & No                             & 37.823        & 41.997          & 49.577         \\
\rowcolor[HTML]{CBCEFB} 
GLUCOSE                          & GLUCOSE           & No                             & 42.51         & 47.186          & 53.856         \\
\rowcolor[HTML]{CBCEFB} 
ATOMIC 2020, ConceptNet, GLUCOSE & GLUCOSE           & Subject                        & 80.879        & 82.014          & 85.664         \\
\rowcolor[HTML]{CBCEFB} 
ConceptNet, GLUCOSE              & GLUCOSE           & Subject                        & 81.349        & 82.775          & 85.926         \\
\rowcolor[HTML]{CBCEFB} 
GLUCOSE                          & GLUCOSE           & Subject                        & 80.928        & 82.076          & 85.681         \\
\rowcolor[HTML]{CBCEFB} 
ATOMIC 2020, ConceptNet, GLUCOSE & GLUCOSE           & Subject, Specificity, Relation & 85.721        & 87.433          & 90.04          \\
\rowcolor[HTML]{CBCEFB} 
ConceptNet, GLUCOSE              & GLUCOSE           & Subject, Specificity, Relation & 85.72         & 87.518          & 90.034         \\
\rowcolor[HTML]{CBCEFB} 
GLUCOSE                          & GLUCOSE           & Subject, Specificity, Relation & 85.65         & 87.473          & 89.967     \\\hline   
\end{tabular}
\caption{Here we present the results of our \hyperref[def:jointcommonsenseinference]{joint inference} tests. We color code sets of rows as testing run on \textcolor{green}{ATOMIC 2020}, \textcolor{CadetBlue}{GLUCOSE}, and \textcolor{Goldenrod}{ConceptNet}. The Training Set(s) column contains the knowledge bases that were used to train the model. The Test set column contains which knowledge base test set was used to evaluate the models. The Hint column represents the Hints that were given to the model during testing. Overall, we can see that with hinting on the test set (i.e., hinting the subject or the subject and the relation type), the addition of knowledge bases for inference does not improve nor degrade substantially the performance. To view this in the table, we can compare the rows in which the Hint column is either “Subject” or “Subject, Relation, Specificity”. Additionally, we can see that without hinting on the test set (i.e., rows that Hint is “No”), the addition of knowledge bases for inference tends to decrease the performance.}
\label{tab:jointinference1}

\end{table}

\begin{table}[]
\resizebox{\textwidth}{!}{

\begin{tabular}{|l|l|l|l|l|l|l|l|l|l|}
\hline
\multicolumn{1}{|c|}{\textbf{Model}}                                                  & \cellcolor[HTML]{9AFF99}\textbf{Acceptable} & \cellcolor[HTML]{9AFF99}\textbf{\begin{tabular}[c]{@{}l@{}}Contextually \\ Acceptable\end{tabular}} & \cellcolor[HTML]{9AFF99}\textbf{\begin{tabular}[c]{@{}l@{}}Alignment \\ Acceptable\end{tabular}} & \cellcolor[HTML]{FFFFC7}\textbf{Acceptable} & \cellcolor[HTML]{FFFFC7}\textbf{\begin{tabular}[c]{@{}l@{}}Contextually \\ Acceptable\end{tabular}} & \cellcolor[HTML]{FFFFC7}\textbf{\begin{tabular}[c]{@{}l@{}}Alignment \\ Acceptable\end{tabular}} & \cellcolor[HTML]{CBCEFB}\textbf{Acceptable} & \cellcolor[HTML]{CBCEFB}\textbf{\begin{tabular}[c]{@{}l@{}}Contextually \\ Acceptable\end{tabular}} & \cellcolor[HTML]{CBCEFB}\textbf{\begin{tabular}[c]{@{}l@{}}Alignment \\ Acceptable\end{tabular}} \\ \hline
\cellcolor[HTML]{9AFF99}ATOMIC 2020 - No Hint                                         & 0.70                                        & 0.66                                                                                                & 0.6                                                                                              & -                                           & -                                                                                                   & -                                                                                                &                                             & -                                                                                                   & -                                                                                                \\ \hline
\cellcolor[HTML]{FFFC9E}ConceptNet - No Hint                                          & -                                           & -                                                                                                   & -                                                                                                & 0.77                                        & \textbf{0.71}                                                                                       & \textbf{0.72}                                                                                    & -                                           & -                                                                                                   & -                                                                                                \\ \hline
\cellcolor[HTML]{CBCEFB}GLUCOSE - No Hint                                             & -                                           & -                                                                                                   & -                                                                                                & -                                           & -                                                                                                   & -                                                                                                & \textbf{0.81}                               & \textbf{0.68}                                                                                       & \textbf{0.8}                                                                                     \\ \hline
\begin{tabular}[c]{@{}l@{}}ATOMIC 2020 - ConceptNet-\\ GLUCOSE - No Hint\end{tabular} & \textbf{0.76}                               & \textbf{0.68}                                                                                       & \textbf{0.63}                                                                                    & \textbf{0.83}                               & 0.7                                                                                                 & 0.65                                                                                             & 0.79                                        & 0.67                                                                                                & 0.59                                                                                             \\ \hline
\begin{tabular}[c]{@{}l@{}}ATOMIC 2020 - ConceptNet-\\  GLUCOSE - Hint\end{tabular}   & 0.71                                        & 0.53                                                                                                & 0.57                                                                                             & 0.77                                        & 0.64                                                                                                & 0.68                                                                                             & 0.77                                        & 0.64                                                                                                & 0.68                                                                                             \\ \hline
\end{tabular}
}
\caption{Results for human annotation of 100 randomly sampled \hyperref[def:assertion]{assertions} from \textcolor{green}{ATOMIC 2020}, \textcolor{CadetBlue}{GLUCOSE}, and \textcolor{Goldenrod}{ConceptNet} test sets and the inferred commonsense from these. We have three sets of three columns Acceptable, Contextually Acceptable, and Alignment Acceptable. Each set of columns is color-coded to represent a knowledge base. Firstly, the Acceptable column is the ratio of whether humans thought that inferred \hyperref[def:assertion]{assertions}, without context, were acceptably commonsense or not. The Contextually Acceptable, column represents the ratio of whether humans thought that inferred \hyperref[def:assertion]{assertions} given the context, were acceptable or not.  Lastly, the Alignment Acceptable column is whether humans thought that the gold standard (from a knowledge base) assertion was correctly matched to the context. We can see that without hinting, the \hyperref[def:jointcommonsenseinference]{joint inference} model (i.e ATOMIC 2020 - ConcetpNet - GLUCOSE - No Hint) improves the acceptability, both with and without context, of \hyperref[def:assertion]{assertions} predicted in the ATOMIC 2020 test set. We can also see that performance does not degrade much in whether it produces \hyperref[def:assertion]{assertions} that are contextually acceptable throughout the test sets. We can see that with hinting, however, the performance is decreased and becomes closer to what the individually trained models can achieve.  This suggests that with hinting, the model tries to channel the knowledge base that we are targetting, and aligns to what we see in the automated metrics.}
\label{tab:mechturk1}
\end{table}

The results for our automated experiments can be found in Table \ref{tab:jointinference1} and from our human experiments in \ref{tab:mechturk1}. From our experiments in this area, we notice the following. Firstly, when used with \hyperref[def:hinting]{hinting}, \hyperref[def:jointcommonsenseinference]{joint inference} \textit{does not} seem to improve the performance of synthetic tests. What this may mean is that \hyperref[def:hinting]{hinting} manages to utilize the format (e.g., relation types) of the knowledge base that it is tapping into for information. Additionally, some of the knowledge sources that we are using have a little overlap (GLUCOSE and ConceptNet had approximately 0.34\% of overlap \cite{mostafazadeh-etal-2020-GLUCOSE}, and ATOMIC 2020 has approximately 9.4\% of overlap with ConceptNet \cite{hwang2020comet}), which means that once \hyperref[def:hinting]{hinting} is utilized to give control signals to the models, this lack of overlap may attribute to why the metrics do not decrease drastically. 
Secondly, without \hyperref[def:hinting]{hinting}, in the automated tests that we run, performance seems to degrade when we add knowledge bases.  Upon further inspection of the results, the reason for seems to be that the model thinks that an \hyperref[def:assertion]{assertion} in the format of another knowledge base (e.g., \hyperref[def:generalinference]{generalized assertion} from GLUCOSE on the ConceptNet test set, relation type from ATOMIC in the GLUCOSE test set) may be more relevant than the test assertion types that we are evaluating. Upon inspecting some inspections of test results, this seems to be the case. We give one such example now, where a model that was trained on all the knowledge bases (ATOMIC 2020, ConceptNet, GLUCOSE) has to predict \hyperref[def:assertion]{assertions} for a test set from ConceptNet:
\begin{quote}
\small
    \textbf{Model Input: }\textbf{Someone I went to school with had a funny laugh.} He would chuckle. Then he would make a whew sound. He had not realized it before. He could not stop his crazy laugh.\\
    
    \textbf{Expected Output:} <general> <subject> person <relation> is/are capable of <object> laugh at joke \\
    
    \textbf{Model Output:} <general> <subject> Someone I went to school with had a funny laugh. <relation> makes others react <object> entertained  \\
    
    \textbf{Model Output with Subject, Relation, Specificity Hint (<general> <subject> person <relation> is/are capable of):} <general> <subject> person <relation> desires <object> laugh at joke 
\end{quote}

The relation “makes others react” is not from ConceptNet, but from ATOMIC 2020. If we \hyperref[def:hint]{hint} the model the subject, relation type, and specificity, it produces a result similar to what we are expecting, where it only defers to using the “desires” relation for the contextual inference. Which, in this case, would not be an incorrect inference. 

Following this, we look at the results for our Mechanical Turk study, can be found in table \ref{tab:mechturk1}. We can see that without hinting, the \hyperref[def:jointcommonsenseinference]{joint inference} model (i.e. ATOMIC 2020 - ConcetpNet - GLUCOSE - No Hint) improves the acceptability, both with and without context, of \hyperref[def:assertion]{assertions} predicted in the ATOMIC 2020 test set. We can also see that performance does not degrade much in whether it produces \hyperref[def:assertion]{assertions} that are contextually acceptable throughout the test sets. We can see that with \hyperref[def:hinting]{hinting}, however, the performance decreases and becomes closer to what the individually trained models can achieve.  This suggests that with \hyperref[def:hinting]{hinting}, the model tries to channel the knowledge base that we are targeting, and this aligns to what we see in the automated metrics. This is reinforced by the test example given previously, and similar examples can be found for the different test sets.

Now, taking these results together, when we use \hyperref[def:hinting]{hinting} and join our multiple knowledge sources to perform this joint inference, we are able to within one model, essentially fit all the knowledge bases, that we are evaluating, at hardly any loss in plausibility/acceptability, or at the cost of automated metrics. This has implications for downstream applications because they no longer require multiple models. With one model and \hyperref[def:hinting]{hinting} we can do what three separate models would do. 

Lastly, we also note that on average our alignment technique has 60\% approval rate for ATOMIC 2020,  68.3\% for ConceptNet, and 69\% for GLUCOSE, which gives us on average 65\% approval for our alignment strategy of using sentence-transformers with the FAISS similarity search.  

\section{Adversarial Language Models}
\subsection{Adversarially training language models}
\begin{figure}[h]
\includegraphics[width=8cm]{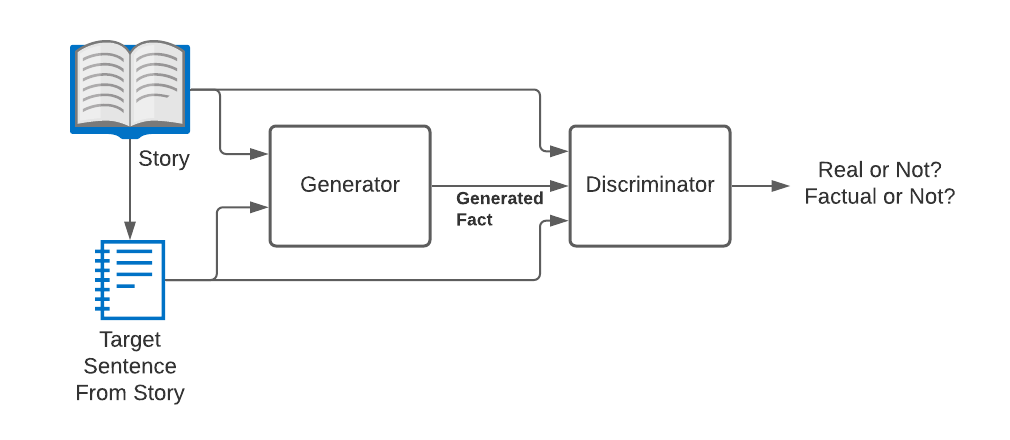}
\caption{Overview of the proposed GAN architecture. A story and a target sentence are fed into the generator, which infers a contextual commonsense fact. This fact, along with the story and target sentence, are passed into a discriminator to determine whether it is from a generator or not and whether it is factual or not.}\label{fig:ganoverview}
\end{figure}

In this work, our main contribution is providing and demonstrating the usefulness of a method for adversarially training language models for the task of \hyperref[def:contextualcommonsenseinference]{contextual commonsense inference}.  In the broader literature of generative adversarial networks (GANs) \cite{goodfellow2014gan}, the adversarial training of models, tends to lead to better results than training each model individually, possibly because of the gradients flowing from the discriminator informing the generator on how to improve. Additionally, the discriminative model can give a measure (usually in a 0-1 range) of how good generations are.   In this work, we propose using the generative language model (i.e., a generator) that we train for contextual commonsense inference and combine it with a discriminative model (i.e., discriminator) whose inputs are the same story and target sentence along with the generator's inference. The discriminator can give a measure of how good the quality of the generated inference is for the given context. This general architecture can be seen in Figure \ref{fig:ganoverview}.\footnote{ One interesting aspect of this formulation, is that it becomes a kind of conditional GAN \cite{mirza2014conditional}, which could reinforce the control signals given in hints.}

To be able to achieve this architecture, we need to be able to connect a generative language model to a regular language model with some additional final layers that produce a score. In our work, we utilize a transformer-based encoder-decoder generative model. Specifically, we use the BART model \cite{lewis-etal-2020-bart}for conditional generation, provided by the Huggingface Transformers library \cite{wolf2019huggingface}. For our discriminator, we utilize a regular BART for sequence classification model also from Huggingface Transformers. However, it is not as simple as conditionally generating, and passing into the discriminator the generated \hyperref[def:assertion]{assertions}. The generative process utilized is a recurrent next token generation. Recall that to pick a next token with this method, a non-differentiable \textit{argmax} operation is used.  This impedes the gradients from being calculated in backpropagation.  The issue becomes even more complex, in that the generation process can utilize beam search to find even better generations, and each beam at the end of each generation step selects a next best token also with an \textit{argmax}.  To address this discontinuity, we utilize an approximation of the \textit{argmax} (i.e., a soft \textit{argmax}) described in the next section and similar to the work described in Section \ref{sec:ganlit}, and perform a dot operation on the scores from this soft-\textit{argmax} with the embeddings from the embedding layer to get an approximate and differentiable input embedding for the discriminator.  Finally, we pick two of the same types of base model (e.g., BART), in order for both the generator and discriminator to share a vocabulary. The reason for sharing the vocabulary is addressed in the next section, however this may not be necessary, and we give an alternative way of being able to “splice” together different models for this task in section \ref{sec:splicing}.

We give some formal notations to describe the GAN framework. Let $G$ be a learnable function (implemented as a Generative Neural model) that can take an input from a domain $X$ and convert it to an output in another domain $Y$, namely $G:X	\rightarrow Y$. That output $Y$ is evaluated by a learnable function $D$ that scores the output $Y$. A \hyperref[def:gan]{generative adversarial network (GAN)} is an interplay between $G$ and $D$, in which $G$ tries to minimize the difference between what it generates, and $D$ tries to maximize its discrimination of fake generations \cite{goodfellow2014gan}. We note that we are not the first to attempt utilizing \hyperref[def:gan]{(GAN)} systems for text generation as seen in Section \ref{sec:ganlit}, but we are the first to apply this system for the task at hand. 

\subsection{Addressing the Discontinuity in Generation}
\label{sec:discontinuity}
Recall that during recurrent conditional language generation, a next token, $N$, is selected by finding the $argmax$ of a $softmax$ of all the vocabulary, after a language  model is given the generated phrase up until step $N-1$. Also recall that an embedding layer is a neural network component that given an index $i$ , returns a row vector, from a vocabulary matrix, that corresponds to $i$. This lookup operation can also be achieved by performing a dot product of a one hot vector that represents the index $i$ and the vocabulary matrix. This essentially scales every row in the matrix by the corresponding vector component and sums all the vectors. In the case of a one-hot vector, it scales all but one vector to zero, therefore leaving only the desired $i$ at the end of the summation. 
\begin{figure}[h]
\begin{minipage}{\linewidth}
\centering
\includegraphics[width=\textwidth]{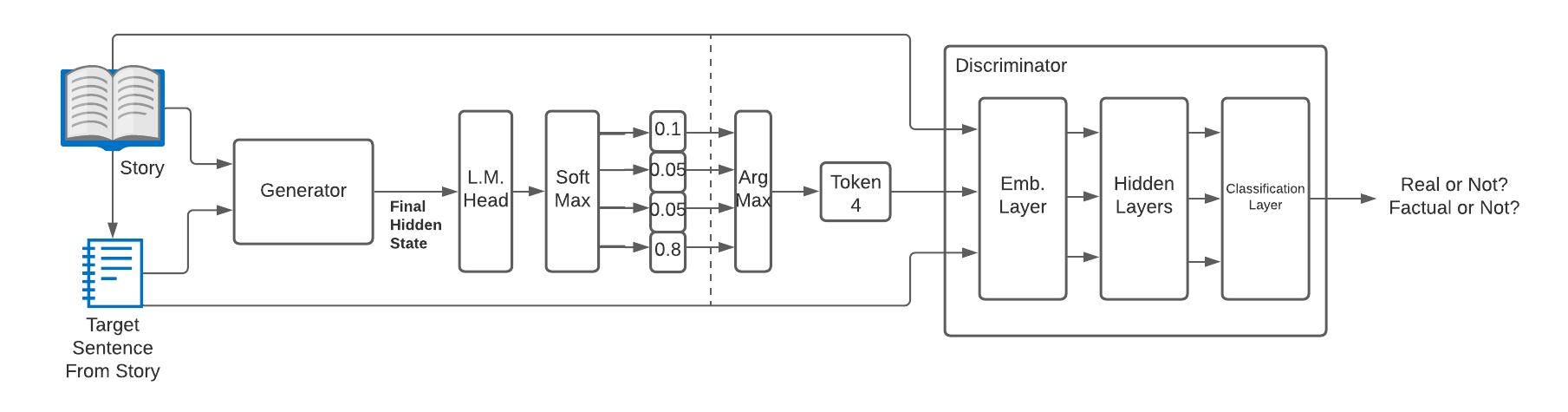}
\caption{We visualize an example that shows the discontinuity when combining a generative language model with a discriminative language model. The dashed line represents where the gradients are discontinued because of the non-differentiable \textit{argmax} operation.}\label{fig:discontinuity}
\end{minipage}

\begin{minipage}{\linewidth}
\centering
\includegraphics[width=\textwidth]{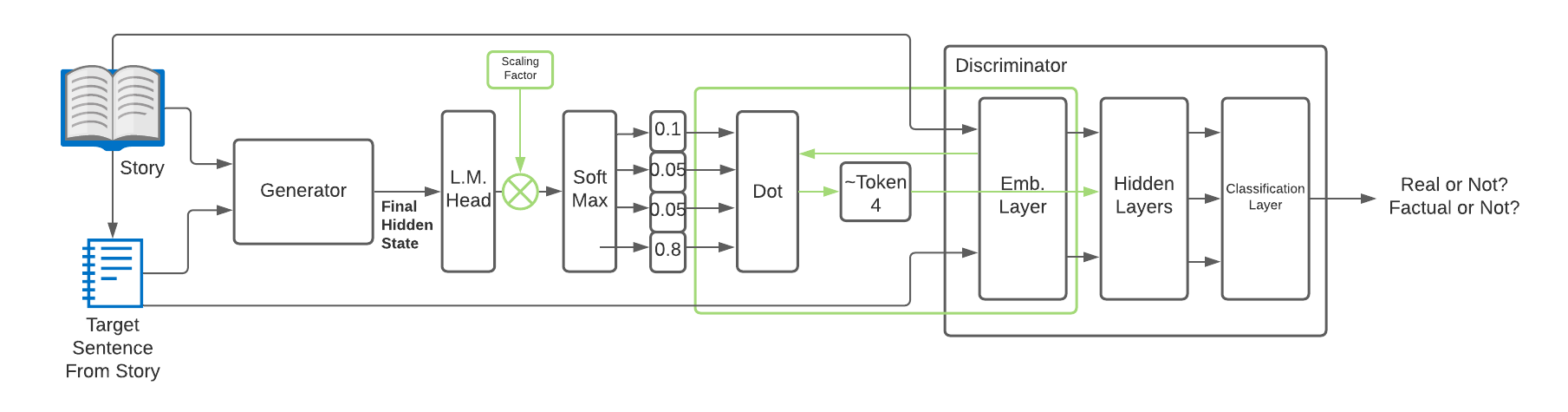}
\caption{We visualize an example that shows how we address the discontinuity by replacing the non-differentiable \textit{argmax} with a dot product between the softmax and the embedding layer matrix. Additionally we highlight where the scaling factor is inserted to make the approximation more accurate. We mark our approach in \textcolor{LimeGreen}{green}.}\label{fig:continuity}
\end{minipage}
\end{figure}

Now, to maintain the gradients, we need to connect the output of our generative model, which is the softmax, to the embedding layer of our discriminator model so that it can be input, scored, and backpropagated correctly. To do this, we simply replace the aforementioned one-hot vector that represents our index, with the softmax that the generative model produces at a given generation step, and perform a dot product of this softmax with the embedding matrix. This method is approximate, given that there may be noise from other non-zero elements in the softmax, and  the top element is not an exact 1. To somewhat remedy this approximation, we can multiply the input of the softmax by a certain factor to essentially give a more approximate one-hot vector.  However, this factor cannot be very large, because it may cause instability during backpropagation. In our work, we use a scaling factor of 1, as this seems to be accurate enough. We repeat this approximation for every generation step, and are left with a list of input embeddings for the discriminator that represent the output of the generator with usable gradients. Since we are using the same vocabulary, we can verify how accurate the output is, by training a K-Nearest Neighbors system, and finding the K=1 neighbor of the output of the softmax and embedding matrix dot product against the embedding matrix. We can use this test to determine an appropriate factor for scaling the softmax. Altogether, this approximation permits us to train our two language models adversarially by having gradients flow from the discriminator to the generator. We note that there is work that utilizes a similar simplification to permit gradient flow in an adversarial system in Section \ref{sec:ganlit}.           
\subsection{Addressing Different Generation Types}
The aforementioned approximation for the discontinuity, as we described it, can be utilized for greedy selection of the next-token (i.e., we always pick the maximal one from the final softmax). We can also apply this technique to beam-search generation, and at every point in constructing the top scored beam, we utilize the softmax of the maximal scoring beam, essentially simplifying the problem back down to a greedy generation-like formulation. In this work however, we do not explore top-k generation, top-p generation, nor sampling during generation.  Top-k and top-p generation can be seen as masking out with zeros, tokens that do not meet a certain criteria. Sampling is more complicated.  To use our approximation with sampling, we would need to model the sampling function at every generation step with something like a recurrent neural network. We leave this line of research for future work.

\subsection{Splicing different models}
\label{sec:splicing}

We come back to address the issue of having to utilize models that have the same vocabulary.  The reason for this is that the soft argmax operation matches in matrix multiplication dimensions between models.  We now give an alternative, although unexplored, option. Given that a generative model will produce a softmax vector of vocabulary size $V$, and we have another model that has a different vocabulary size of $M$, we can train decoding layers that can convert the output tokens of the generative model, into corresponding tokens from the discriminative model. However, this conversion layer would need to be trained beforehand, and may need to be frozen during the adversarial training, otherwise it would be a disconnect between the two models, and the input given to the discriminator may be corrupted. To train this conversion layer beforehand, one could use as a ground truth, the results that a tokenizer from model B, with the vocabulary size of $M$, would use as the targets in a cross entropy loss, and the results that a tokenizer from model A, with the vocabulary size of $V$, uses as the input to the layer.          
\subsection{Factuality in the Discriminator}
Given that we can now adversarially train our models, we explore enhancing the discriminator with some way to determine factuality. We take a simple approach that in addition to the normal discriminator training objective (i.e., the discriminator is given a batch of generated text and a batch of real text and evaluated whether it inferred this correctly), we add a confounder loss.  Our additional confounder loss is based on the confounder loss by \cite{li2016commonsense}, in that we shuffle around the subjects and objects and expect our model to determine that when shuffled objects are false.  Since our generated outputs are structured (i.e., we have symbol tokens that delimit the different parts of \hyperref[def:assertion]{assertions}), we can do this shuffling easily. Although shuffling may incur in some false negatives (we may have a shuffled configuration that is factually correct), since we supply the story and target sentence, we expect the discriminator to be able to discern this correctly. We believe that we could also apply the max-margin loss utilized to great effectiveness by other language GAN literature \cite{ponti-etal-2018-adversarial,colon-hernandez-etal-2021-retrogan}, although we leave this for future work.

\subsection{Experimental Setup}
For our GAN experiments, we had the following setup. We built a \hyperref[def:jointcommonsenseinference]{joint inference} dataset using the procedure given in section \ref{sec:jointhinting}, and we augmented it with hinting.  Hinting was done in the same manner as in Section \ref{sec:whatishinting}, by sampling a binomial distribution ($p=0.5$) and if the sample is true we provide a \hyperref[def:hint]{hint} by randomly sampling parts of the target \hyperref[def:assertion]{assertion}. This dataset is composed of 1,479,811 training examples, and 30,183 testing examples. We fed this data to a model with the adversarial setup described in the previous sections. The generator model utilized was the BART-base for conditional generation, and the discriminator model was a BART-base model for sequence classification. For the sake of time, we run our model on only 100,000 examples, with a batch size of 32, on 3xNVIDIA A6000 for 3 epochs. In addition to this, to see the effects of the adversarial formulation and of the confounder loss, we train a model without the adversarial approach that we propose (a separated Generator and Discriminator with the confounder loss), and an adversarial model without the confounder loss to be able to gauge the effects of it. We train 4 random seeds for each of these 3 conditions. Additionally, to test the performance of the discriminator, we used the alignment technique from Section \ref{sec:jointhinting} to align the ConceptNet test set of \cite{li2016commonsense} to the ROCStories corpus. We then passed the story, target sentence and the test assertion into the discriminator to determine whether it was true or not. We used a threshold of 0.5 to determine whether an assertion was marked as true (1) or false (0). 
% We run an additional user study, which consists of randomly sampling 200 entries from our test data, and running the same setup as the study described in section \ref{sec:jointhinting}, we ask a pair of human annotators to determine whether the generated \hyperref[def:assertion]{assertions} are valid, and whether they are valid for the given context; the same setup as for our joint inference.  

\subsection{Effects of adversarial training}
% Please add the following required packages to your document preamble:
% \usepackage[table,xcdraw]{xcolor}
% If you use beamer only pass "xcolor=table" option, i.e. \documentclass[xcolor=table]{beamer}
% Please add the following required packages to your document preamble:
% \usepackage[table,xcdraw]{xcolor}
% If you use beamer only pass "xcolor=table" option, i.e. \documentclass[xcolor=table]{beamer}
% Please add the following required packages to your document preamble:
% \usepackage[table,xcdraw]{xcolor}
% If you use beamer only pass "xcolor=table" option, i.e. \documentclass[xcolor=table]{beamer}
\begin{table}[h]
\resizebox{\textwidth}{!}{
% Please add the following required packages to your document preamble:
% \usepackage[table,xcdraw]{xcolor}
% If you use beamer only pass "xcolor=table" option, i.e. \documentclass[xcolor=table]{beamer}
\begin{tabular}{|l|l|l|l|l|l|l|l|}
\hline
\rowcolor[HTML]{B0B3B2} 
\textbf{Model}                                           & \textbf{ROUGE1} & \textbf{ROUGE2} & \textbf{ROUGEL} & \textbf{ROUGELSUM} & \textbf{BLEU}   & \textbf{METEOR} & \textbf{Accuracy of Discriminator} \\ \hline
\cellcolor[HTML]{D4D4D4}\textbf{+ADVERSARIAL+CONFOUNDER} & 43.656          & \textbf{10.544} & 40.380          & 40.379             & 31.335          & 61.683          & 0.690                              \\ \hline
\cellcolor[HTML]{D4D4D4}\textbf{+ADVERSARIAL-CONFOUNDER} & \textbf{43.747} & 10.559          & \textbf{40.530} & \textbf{40.531}    & 31.279          & 61.623          & 0.481                              \\ \hline
\cellcolor[HTML]{D4D4D4}\textbf{-ADVERSARIAL+CONFOUNDER} & 43.715          & \textbf{10.680} & 40.292          & 40.292             & \textbf{31.470} & \textbf{61.776} & \textbf{0.794}                     \\ \hline
\end{tabular}
}
\caption{We present the results of the adversarial test with ablations. We can see that the Adversarial models tend to have improved recall (ROUGE) scores, but lower precision (BLEU/METEOR) scores and lower accuracy on classifying a ConceptNet test set of \hyperref[def:assertion]{assertions}. The adversarial model with everything (+ADVERSARIAL+CONFOUNDER) strikes a balance of the benefits of precision and recall that the non-adversarial, non-confounder loss model give respectively.}
\label{tab:mechturk2}
\end{table}
After running automated tests, we see some mixed results between the three conditions (Adversarial+Confounder, -Adversarial+Confounder, Adversarial-Confounder). These results are in Table \ref{tab:mechturk2}. We can see that the Adversarial models tend to have improved recall (ROUGE) scores, but lower precision (BLEU/METEOR) scores and lower accuracy on classifying a ConceptNet test set of \hyperref[def:assertion]{assertions}. The adversarial model with everything (+ADVERSARIAL+CONFOUNDER) strikes a balance of the benefits of precision and accuracy, and recall that the non-adversarial, non-confounder loss model give respectively. Some possible causes for these mixed results  may be that our approach may be too naive, and possibly an improved \hyperref[def:gan]{GAN} formulation such as the Wasserstein GAN \cite{pmlr-v70-arjovsky17a} used in \cite{press2017language} may help our results, our approximation to connect the generator and discriminator may be too naive and may need a more complex approach such as utilizing a recurrent neural network during the generation steps to encode them then decode them into the discriminator.

\section{Contributions \& Takeaways}
In this work we have presented three things: a method for controlling \hyperref[def:contextualcommonsenseinference]{contextual commonsense inference} called hinting, a method for combining multiple \hyperref[def:knowledgegraph]{knowledge graphs }for joint \hyperref[def:contextualcommonsenseinference]{contextual commonsense inference}, and an adversarial and non-adversarial model trained with these techniques. Taken altogether, we can obtain one model that is capable of inferring on a topic given by a hint, the inference can be performed on any \textcolor{Emerald}{subject}, \textcolor{Peach} {object}, or a  \textcolor{Orchid}{relation}, and specificity from source knowledge bases, and the model's discriminator is capable of scoring \hyperref[def:assertion]{assertions}. These contributions serve as a baseline to explore the area of \hyperref[def:contextualcommonsenseinference]{contextual commonsense inference}, and leave much room to explore avenues on hinting, joint inference, and adversarial training of transformer-based language models. 

%\section{}\label{s1}

%\subsection{}\label{s1.1}

%\begin{figure}[t]
%\includegraphics{}
%\caption{Figure caption.}\label{f1}
%\end{figure}

%\begin{table*}
%\caption{} \label{t1}
%\begin{tabular}{lll}
%\hline
%&&\\
%&&\\
%\hline
%\end{tabular}
%\end{table*}

%%%%%%%%%%% The bibliography starts:

%%%%%%%%%%%%%%%%%%%%%%%%%%%%%%%%%%%%%%%%%%%%%%%%%%%%%%%%%%%%%
%%                  The Bibliography                       %%
%%                                                         %%
%%  ios1.bst will be used to                               %%
%%  create a .BBL file for submission.                     %%
%%                                                         %%
%%                                                         %%
%%  Note that the displayed Bibliography will not          %%
%%  necessarily be rendered by Latex exactly as specified  %%
%%  in the online Instructions for Authors.                %%
%%                                                         %%
%%%%%%%%%%%%%%%%%%%%%%%%%%%%%%%%%%%%%%%%%%%%%%%%%%%%%%%%%%%%%

% \nocite{*}
% if your bibliography is in bibtex format, use those commands:
\bibliographystyle{ios1}           % Style BST file.
\bibliography{bibliography}        % Bibliography file (usually '*.bib')
\pagebreak
\appendix
\section{Definitions}
\begin{description}
    \item[Contextual/Discourse-Aware Commonsense Inference] Task in which, given a textual context (e.g., story) and a selected sentence from that context, a model must infer a coherent and contextual commonsense assertion.\label{def:contextualcommonsenseinference}
    \item[Assertion] Used interchangeably with a fact, it is a tuple that consists of a subject, relation, object, and generality and represents a fact. \label{def:assertion}
    \item[Sentence-Level Commonsense Inference]Generation of a commonsense assertion utilizing, at most, a sentence as context.\label{def:sentencecommonsenseinference} 
    \item[Story specific commonsense assertion inference] Commonsense assertion templates that are instanced by elements from a story.\label{def:storyspecificinference} 
    \item[General commonsense assertion inference]Commonsense assertion templates that are not instanced, but are derived from a story. \label{def:generalinference} 
    \item[Specificity] Whether an assertion's content is about entities in the aligned story (i.e., given context), or if it is a generalized version of an assertion.\label{def:specificity} 
    \item[Joint Commonsense Inference] To infer commonsense knowledge \hyperref[def:assertion]{assertions} by leveraging knowledge from multiple knowledge bases.\label{def:jointcommonsenseinference} 
    \item[Hint]The part(s) of an assertion that a model has to predict, along with special identifiers for these parts, wrapped within parenthesis characters, that are passed to a model to infer a commonsense assertion.\label{def:hint} 
    \item[Hinting]Proposed technique that trains a \hyperref[def:contextualcommonsenseinference]{contextual commonsense inference} model that can be guided on its inference using \hyperref[def:hint]{hints}.\label{def:hinting} 
    \item[Knowledge Graph] A \hyperref[def:knowledgegraph]{knowledge graph }(used somewhat interchangeably with knowledge base, although they are different concepts) is defined as “a graph of data intended to accumulate and convey knowledge of the real world, whose nodes represent entities of interest and whose edges represent relations between these entities” \cite{hogan2020knowledge}.  Formally, a \hyperref[def:knowledgegraph]{knowledge graph }is a set of tuples that represents nodes and edges between these nodes.  Let us define a set of vertices (which we will refer to as concepts) as $V$, a set of edges as $E$ (which we will refer to as assertions as per Speer and Havasi \cite{speer2012representing}), and a set of labels $L$ (which we will refer to as relations). A \hyperref[def:knowledgegraph]{knowledge graph }is a tuple $G:=(V,E,L)$. We use the formal definitions found in Appendix B of \cite{hogan2020knowledge}. The set of edges ($E$) or assertions is composed of tuples $E \subseteq V\times L\times V$ which are seen as a subject (a concept), a relation (a label), and object (another concept) respectively (e.g., $(subject,relation,object)$).  These edges in some cases can have weights to represent the strength of the assertion, and in this work they additionally have a parameter of \hyperref[def:specificity]{\textit{specificity}} (whether the assertion's content is about entities in the aligned story, or if it is a generalized version of an assertion). Broadly speaking, \hyperref[def:knowledgegraph]{knowledge graphs }(KGs) are a collection of tuples that represent things that should be true within the knowledge of the world that we are representing.\label{def:knowledgegraph}  
    \item[Generative Adversarial Network (GAN)] Adversarial system in which a Generator model produces inferences that are scored by a Discriminator system. The Discriminator provides feedback to the Generator on how to improve at the same time that it tries to improve itself on Discriminating Generated and Real data. A more detailed explanation can be found in \cite{goodfellow2014gan}.\label{def:gan}  
    
\end{description}

% or include bibliography directly:
%\begin{thebibliography}{0}
%\bibitem{r1} F. Author, Information about cited object.
%
%\bibitem{r2} S. Author and T. Author, Information about cited object.
%\end{thebibliography}

\end{document}